\documentclass[runningheads]{llncs}
\usepackage{graphicx}

\usepackage{epsfig}
\usepackage{subcaption}
\usepackage{calc}
\usepackage{amssymb}
\usepackage{amstext}
\usepackage{amsmath}
\usepackage{multicol}
\usepackage{pslatex}
\usepackage{apalike}
\usepackage{url}
\usepackage{float}

\begin{document}
\title{Alternative Data Augmentation for Industrial Monitoring using Adversarial Learning}

\author{Silvan Mertes\inst{1} \and
Andreas Margraf\inst{2} \and
Steffen Geinitz\inst{2} \and
Elisabeth André\inst{1}
}
\authorrunning{Mertes et al.}
\institute{University of Augsburg, Universitätsstraße 1, 86159 Augsburg, Germany
\email{\{silvan.mertes,elisabeth.andre\}@informatik.uni-augsburg.de}\\ \and
Fraunhofer IGCV, Am Technologiezentrum 2, 86159 Augsburg, Germany\\
\email{\{andreas.margraf, steffen.geinitz\}@igcv.fraunhofer.de}}

\maketitle              % typeset the header of the contribution
\begin{abstract}
Visual inspection software has become a key factor in the manufacturing industry for quality control and process monitoring. Semantic segmentation models have gained importance since they allow for more precise examination. These models, however, require large image datasets in order to achieve a fair accuracy level. In some cases, training data is sparse or lacks of sufficient annotation, a fact that especially applies to highly specialized production environments. Data augmentation represents a common strategy to extend the dataset. Still, it only varies the image within a narrow range. In this article, a novel strategy is proposed to augment small image datasets. The approach is applied to surface monitoring of carbon fibers, a specific industry use case. We apply two different methods to create binary labels: a problem-tailored trigonometric function and a WGAN model. Afterwards, the labels are translated into color images using pix2pix and used to train a U-Net. The results suggest that the trigonometric function is superior to the WGAN model. However, a precise examination of the resulting images indicate that WGAN and image-to-image translation achieve good segmentation results and only deviate to a small degree from traditional data augmentation. In summary, this study examines an industry application of data synthesization using generative adversarial networks and explores its potential for monitoring systems of production environments.
\end{abstract}

\section{Introduction}\label{sec:introduction}

\subsection{Motivation}
\noindent Visual inspection respresents a wide spread methodology in industrial quality control which is usually employed in mass production to ensure quality standards. With recent progress in deep learning research, the focus of computer vision shifted from image processing filters to neural network architecture selection and design. The increasing level of automation and digitalisation in the manufacturing industry has drawn attention to sensing technology and camera sensors in particular. The field of online process monitoring (OPM) primarily deals with imaging technology to detect changes, faults or potential anomalies in continuous production environments. Of course, the actual benefit of measurement technology depends on its level of automation. Therefore, intelligent image processing is a key feature of monitoring systems.

In recent years, machine learning algorithms have progressively overtaken filter based image processing, a fact that has been discussed in relevant publications ~\cite{Cavigelli2017,McCann2017}. In the context of computer vision, convolutional neural networks (CNNs) have proven to be superior because they generalize better. Given large training and sample data, they are flexible across domains and therefore can be applied to very different kinds of applications ~\cite{simonyan2014very,he2016deep}.

\subsection{Pushing the Limits of Image Segmentation}
Highly specialized industries are often confronted with incomplete or insufficient data. With increasing effort spent on data collection and preparation - tasks that require time and skilled personel - deep learning models become inefficient, expensive and therefore unattractive. ML solutions only serve the cause if they add to productivity. This article discusses concepts that augment scarce datasets and allow semantic segmentation models to achieve a high accuracy when trained on this data.   

Research work has been conducted in the field of data preparation, cleaning and augmentation using e. g. algorithms for interpolation, smoothing, simple transformation or filtering \cite{shorten2019survey}. Although these methods help to create the precondition for successful model development, they quickly appear constrained in their variation space. This article takes a closer look to a branch of research exploring the representation of artificially generated data based on real world blueprints, an approach also denoted \textit{synthetic data generation}.
% ==================================================
% Novel Strategies for Data Augmentation ( Before: Toward better Data Augmentation)
% ==================================================
In this article we consider semantic segmentation on carbon fiber textiles with unique surface structures and heterogeneous anomalies. Earlier publications have shown that image processing based on conventional filters such as edge, contour, threshold or Fourier transformation do not cover anomalies with the desired accuracy. In this respect, we offer a CNN-based method using two differnt models for adverarial learning to augment the data. Thus, our DA methods are based on triginometric functions, \textit{Wasserstein Generative Adversarial Networks} (WGANs) and \textit{pix2pix} image-to-image translation. As an overall goal, we pursue this approach to improve the reliability of defect detection and reduce the costs of model training by automating a considerable portion of the data preparation tasks.

One of the main advantages of CNNs is their ability to classify images, i.e. to predict which category an image, object within the image or single pixel belong to. This dependency is usually determined on the basis of the "euclidian distance", a widely used similarity score. The latest generation of CNNs requires large amounts of samples and data to achieve reasonable results. Manual annotation cannot be fully automated to this date, but still constitutes a time consuming task and - since it is performed by people - is prone to error over longer periods of time. Regular DA has been a popular and obviously simple tool to increase the number of training samples. DA methods include rotation, scaling, blurring and similiar transformations. These functions adjust image and label pairs carefully and to a limited extent but do not exploit the full potential that is offered by \textit{Generative Adversarial Networks} (GANs) and image-to-image translation.

GANs are designed to output `new' images which appear to be as realistic as possible, so that they are indistinguishable from real-world photographs. Image-to-image translation as performed by \textit{pix2pix} improves the ability of the original GANs to transfer images from one domain to another. However, input data for GAN training should be carefully collected or created with the expected application domain in mind. In general, there are two possible approaches to provide the input. One very obvious approach would be to gather data which already contain the desired image-to-image translation information. This could be accomplished by e. g. acquiring images from different angles using the same camera or by moving the target around. This has been examined for the fashion industry in recent publications ~\cite{Liu2019AIFashion,cui2018fashiongan}.

In the manufacturing industry such intensive interfering with the process is undesirable because it may cause time delays or unscheduled interruptions. Therefore, getting access to the target from more than one angle may not be possible. As an alternative approach, one can use simluated data or approximate image-to-image translation by using a specifically designed algorithm. Of course, the second suggestion can only be applied to smaller sets of data in order to keep the workflow efficient. 

In this article, image-to-image translation is used to translate randomly generated binary labels to images. The approach allows for artificially creating label and image pairs that are actually new and serve as training samples for semantic segmentation models. For the generation of new labels, we propose two distinct concepts: the first approach is based on a function precisely tailored to the application and was already introduced by Mertes et al. \cite{mertes2020}. The second one is presented the first time in this paper and uses a WGAN model trained on the original binary labels in order to allow the generation of synthetic labels. In both cases, binary label images are generated and used for further processing, i.e. for generating image/label pairs by using an image-to-image translation system. Designing problem-specific functions requires good domain understanding and further effort to model an algorithm and tune it to adjust to the given problem. The resulting function is transparent and human-readible which allows for better debugging and testing. However, the proposed WGAN-based approach requires less manual effort and automates a considerable part of the overall process.

All in all, we present a novel approach, to augment image data for semenatic segmentation networks by applying image-to-image translation with both, a domain-specific mathematical model and an approach entirely based on generative models. We test both approaches based on images containing carbon fiber surface defects and discuss the results.

\subsection{Structure}
The structure of this article is divided into four main sections: 
at first, we provide an overview on related work and existing technology in section \ref{sec:relatedwork}. The subsequent section present the two approaches (section \ref{sec:approach}) followed by the experimental setup (section \ref{sec:experiments}). We then discuss relevant results and critically reflect them in respect of related concepts in section \ref{sec:discussion}. The final section covers concluding statements based on our findings and provides an outlook on future work as can be seen in section \ref{sec:conclusion}.

\section{Related Work}\label{sec:relatedwork}
This section lists and discusses previously published papers in related research areas which include Machine Learning (ML), Artificial Neural Networks (ANN), Computer Vision (CV), GANs, OPM and Organic Computing (OC) as well as publications that contributed to the research presented in this article.
% ===
% OPM
% ===
The identification of anomalies on carbon fibers, e.g. the misalignment on textile surfaces, has been discussed in various publications \cite{geinitz2016detection,geinitz2016linescan}. In the same context, Margraf et al. examined the self-adaptation of image processing filters using organic computing paradigms \cite{Margraf2017}.
% =================
% Organic Computing
% =================
Methodologies for automated algorithm selection and filter pipelines have been discussed by Stein et al. \cite{Stein2018}. A partly self-adaptive algorithm for data analysis based on carbon fiber monitoring was introduced by Margraf et al. \cite{margraf2020}.

% ===
% CNN
% ===
The domain of artifical neural networks took large steps forward when AlexNet \cite{krizhevsky2012imagenet}, GoogleNet \cite{szegedy2015going} and VGGNet \cite{simonyan2014very} were introduced for the classification of large image sets. Several publications address industrial monitoring applications: Masci et al. used CNN for classification of steel defects, and Soukup et al. used CNN for photometric stereoscopic images ~\cite{masci2012steel,soukup2014convolutional}. A region proposal network for real-time object detection was presented by Ren et al., while Ferguson et al. used CNNs and transfer learning to detect X-ray image defects ~\cite{NIPS2015_5638,ferguson2018detection}. Furthermore, the use of CNNs for industrial surface anomaly inspection was explored by Staar et al. \cite{staar2019anomaly}.
The first ones to introduce pixel-based segmentation was Long et al. \cite{Long_2015_CVPR}, while Schlegl et al. published a work in which GANs for marker detection were used for unsupervised anomaly detection \cite{schlegl2017unsupervised}. A survey exploring GANs for anomaly detection was presented by Di Mattia et al. \cite{mattia2019survey}.
Methods for color translation between different photographic contexts were discussed in various related publications ~\cite{zhang2016colorful,xie2015holistically}.
Translation of completely different image domains was first achieved by Isola et al., who presented the \textit{pix2pix} architecture \cite{isola2017image}. The \textit{pix2pix} architecture was the first to allow the projection of various image domains such as edge objects or label images to colored photographs. 
A huge problem when dealing with pixel-based segmentation tasks is a big gap between a small foreground and a relatively large background. This is especially reflected in the case of carbon fiber suface images. Ronneberger et al. confirmed, that multi-channel feature maps prove to be more useful in these tasks \cite{ronneberger2015u}. It should be noted that this architecture makes heavy use of data enhancement, while taking only little input data. However, its capabilities are limited when dealing with very small training sets.
Multiple approaches to use GANs for the specific purpose of data augmentation were presented in different works. Frid-Adar et al. and Mariani et al. were able to apply GANs to classification tasks, while Choi et al. presented an approach that makes use of image-to-image translation networks for semantic segmentation tasks by transforming labeled data to related image domains so that the original label still fits to the artificially created image ~\cite{frid2018synthetic,mariani2018bagan,choi2019self}. Huang et al. utilized image-to-image networks for shared segmentation tasks by using multiple image domains for the training process \cite{huang2018auggan}.
The authors of this paper could not identify any publication dedicated to the use of GANs for generating entirely new image and label pairs for enhanced training datasets in the context of semantic image segmentation.

% =================
% More GAN Papers
% =================
An early application of generative models was presented by Pathak et al. for image inpainting  to reconstruct a region within an image that adapts well to its surroundings \cite{pathak2016context}.
Shrivastava et al. proposed an approach denoted Simulated+Unsupervised (S+U) learning which allows to improve the output of a simulator based on a GAN network \cite{shrivastava2017learning}.
Furthermore, \textit{StarGAN} was introduced by Choi et al. to improve the quality of translated images by providing a more scalable method for image-to-image translation to an arbitrary domain  \cite{choi2018stargan}.

% ==================================
% WGAN
% ==================================
The WGAN architecture constitutes a variant of generative models. It was introduced to increase learning stability and is less prone to mode collapse. Its structure ressembles the general GAN approach, except for its discriminator that learns on the basis of the Wasserstein distance by approximating a 1-Lipschitz function \cite{arjovsky2017wasserstein}. Later, Arjosvky et al. made further adjustments to the weight clipping which improved the WGAN behaviour to better handle hyperparameter tuning \cite{gulrajani2017improved}. In addition, a comparative study increased the understanding of GAN behaviour regarding training stability and model saturation \cite{arjovsky2017towards}.
% ==================================
% Neurevoluation and Metaheuristic
% ==================================
The authors of this article are aware that metaheuristics for hyperparameter optimization, e. g. swarm intelligence \cite{strumberger2019convolutional} exist. However, this field of research is not subject to this research article. 

\section{Approach}\label{sec:approach}
The following sections explain the concepts that are introduced in this paper. The main idea of our approach is that we are enhancing datasets with augmented data by artificially modeling label images and after that convert them into real image data. By doing so, we get image/annotation pairs that are needed for the training of neural networks for semantic segmentation tasks. To create new label data, we propose two different methods. The first method, that we already introduced in \cite{mertes2020},
is an algorithm specifically designed for our particular application at hand, i.e. the defect detection in carbon fiber structures. It is based on a randomized label generator that uses a stochastically parametrized function to build segmentation masks. The second method is a more generic one. It uses a WGAN that is trained on raw segmentation masks. After training, the WGAN is capable of generating new label images that appear similar to the original labels. By applying this concept, we get rid of the engineering overhead that is necessary when using the first method. While the randomized label generator that is used by the first method has to be defined and optimized specifically for every new segmentation task, the WGAN should be able to learn the label structure of new tasks by itself. \\
The labels that are produced by either of the two methods are fed into a \textit{pix2pix} network that was trained on an image-to-image conversion task, i.e. the network was trained to convert label mask images to real images that fit to the respective labels, thus resulting in image/label pairs that can be used to enlarge training datasets. \\
All in all, our approach can be seen as a three-folded system: first, we train a \textit{pix2pix} network on an image-to-image translation task, so that it learns to perform a translation from labels of defect images to their corresponding image data. Second, one of the aforementioned methods is used to generate synthetic label data.  
At last, the synthetic label data is fed into the trained \textit{pix2pix} network, resulting in new training pairs for further machine learning tasks. The following sections explain these steps in more detail.

\begin{figure}
	\includegraphics[width=1\textwidth]{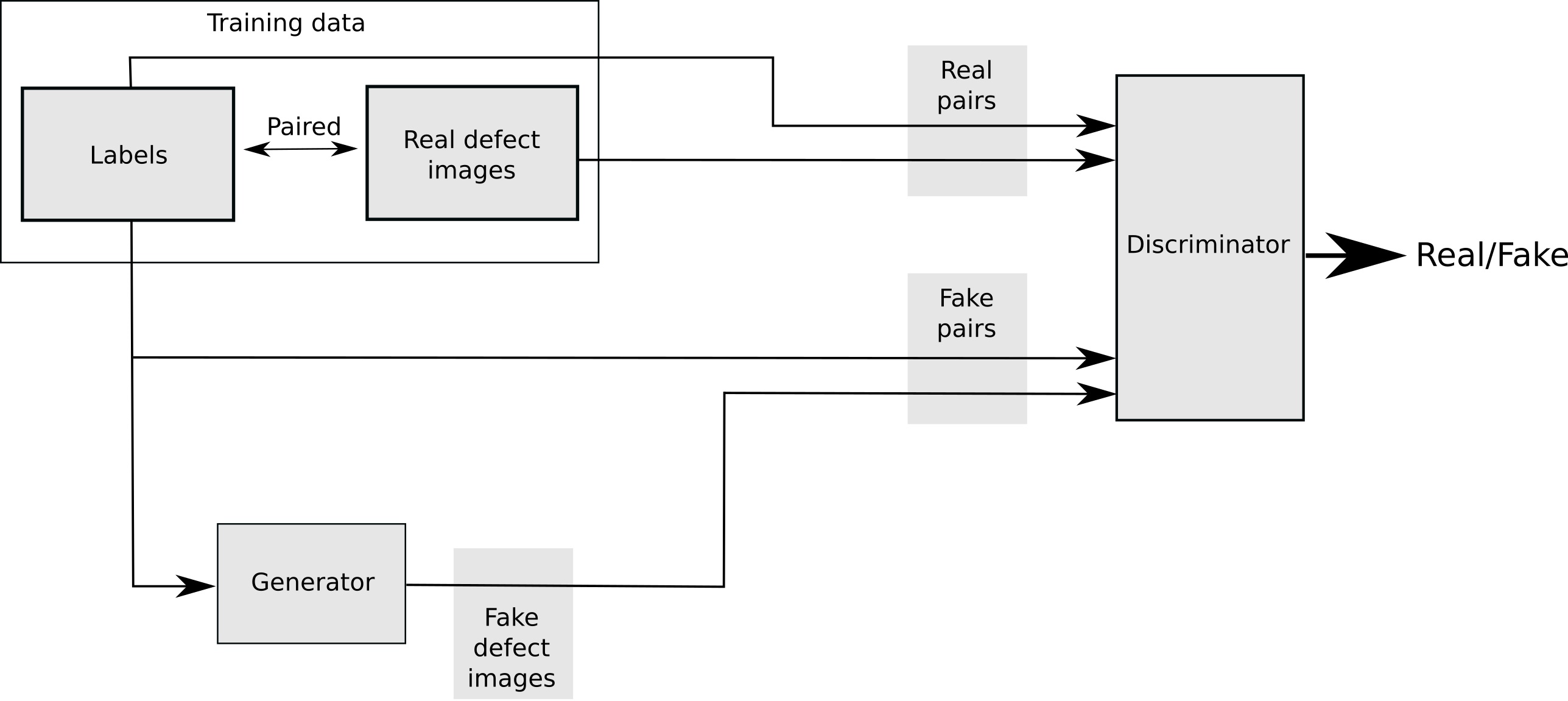}
	\caption{Training of a \textit{pix2pix} network to perform image-to-image translation between labels and defect images (Step 1) \cite{mertes2020}}\label{fig:step_1}
\end{figure}
\subsection{Label-to-Image Model}
In order to convert label masks to corresponding images, we trained a \textit{pix2pix} model. For the training, a dataset of existing real defect images and manually labeled annotation masks was used. The basic scheme of the training process is depicted in Fig. \ref{fig:step_1}. We used the \textit{pix2pix} network architecture that was introduced by \cite{isola2017image}. We adapted the size of the input layer to fit the dimensions of our dataset. Other than that, we did not make any modifications to the originally proposed architecture. 

\subsection{Synthetic Label Generation}
The idea of our approach is to feed new synthetic label data into the \textit{pix2pix} network in order to obtain new pairs of defect images and label masks. As mentioned above, two different methods were applied for the stage of synthesizing new label data.
\subsubsection{Mathematical Modelling of Defects}
\begin{figure}
\centering
\includegraphics[width=0.7\textwidth]{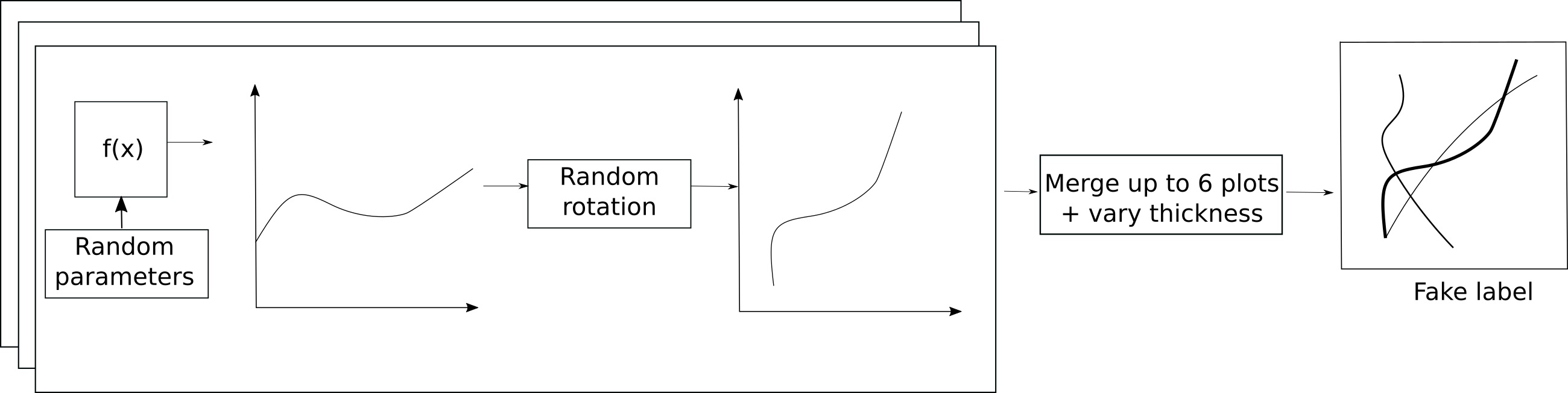}
\caption{Heuristic to generate fake labels using the label generator (Step 2) \cite{mertes2020}}\label{fig:step_2}
\end{figure}

The first method is based on the observation that in many application scenarios label masks have a common structure as proposed by Mertes et al. in their related publication \cite{mertes2020}. It is illustrated in Fig. \ref{fig:step_2}.
The idea behind this first approach to generate fake label masks is to find a mathematical description of those structures for a specific case. In the application scenario that serves as an example for evaluation in this article - the mentioned defect detection on carbon fiber structures - label masks usually appear as mostly straight or curved lines. Those structures can be seen as a combination of multiple graphs with different rotations and varying thickness of the plots. Thus, the mathematical description of a single defect label could be approximated through a trigonometric function. By adding a stochastic factor to such a function, we can plot different graphs that can be considered as new, artificial label masks. For our specific task, we conducted several experiments that showed that the following function $f(x)$ can be used to cover a huge part of carbon fiber defect structures. We denote $f(x)$ as already presented in \cite{mertes2020}:
\begin{multline*}
        f(x) = a_1 \cdot sin(a_2 \cdot x) + a_3 \cdot sin( x) + a_4 \cdot cos(a_5 \cdot x) + a_6\cdot x + a_7\cdot x^2
\end{multline*}
where the parameters $a_n$ are chosen randomly within certain defined intervals. For our specific experiments, we found appropriate intervals by visual expection of carbon fiber defect images. By using those intervals, that are listed in Tab. \ref{tab:function_generator_bounds}, we ensure to cover a wide range of different defect structures. To that end, a sine function was tuned with a rather big amplitude to model the global structure of the label, that is typically shaped in curves. For the structure on a more microperspective level, we used another sine with a much smaller amplitude interval. Aperiodic curvings were modeled by the use of polynomic functions.
As described in \cite{mertes2020}, we randomly set the variables and plotted the resulting graph for every fake label for $x \in [0, w]$ where $w$ represents the width of the sample images. After creating those plots they were rotated randomly. At last, we took a random number of those graphs, randomized the thickness of the resulting lines, and overlapped those graphs to create images of labels with a realistic fiber-like appearance.

The authors are aware of the fact that this method is very specific to the given task at hand. A lot of engineering time and effort has to be spent to find sufficient mathematical models for the respective task. However, similar approaches for defect-modeling have been successfully applied to similiar problems in previous work \cite{haselmann2017supervised}. 

\begin{table}[ht]
\centering
\caption{Parameters for the fake label generator \cite{mertes2020}}
 \begin{tabular}{|c | c | c |}
 \hline
 Parameter & Lower bound & Upper bound \\ [0.5ex] 
 \hline\hline
 $a_1$ & 15 & 30\\
 $a_2$ &  0.02 & 0.03\\
 $a_3$ & 1 & 50\\
 $a_4$ & -0.5 & 0.5\\
 $a_5$ & -0.5 & 0.5\\
 $a_6$ & -0.5 & 0.5\\
 $a_7$ & 0.005 & 0.0095\\
 \hline
\end{tabular}\label{tab:function_generator_bounds}
\end{table}

\subsubsection{Generative Modelling of Defects}

\begin{figure}
    \centering
	\includegraphics[width=0.8\textwidth]{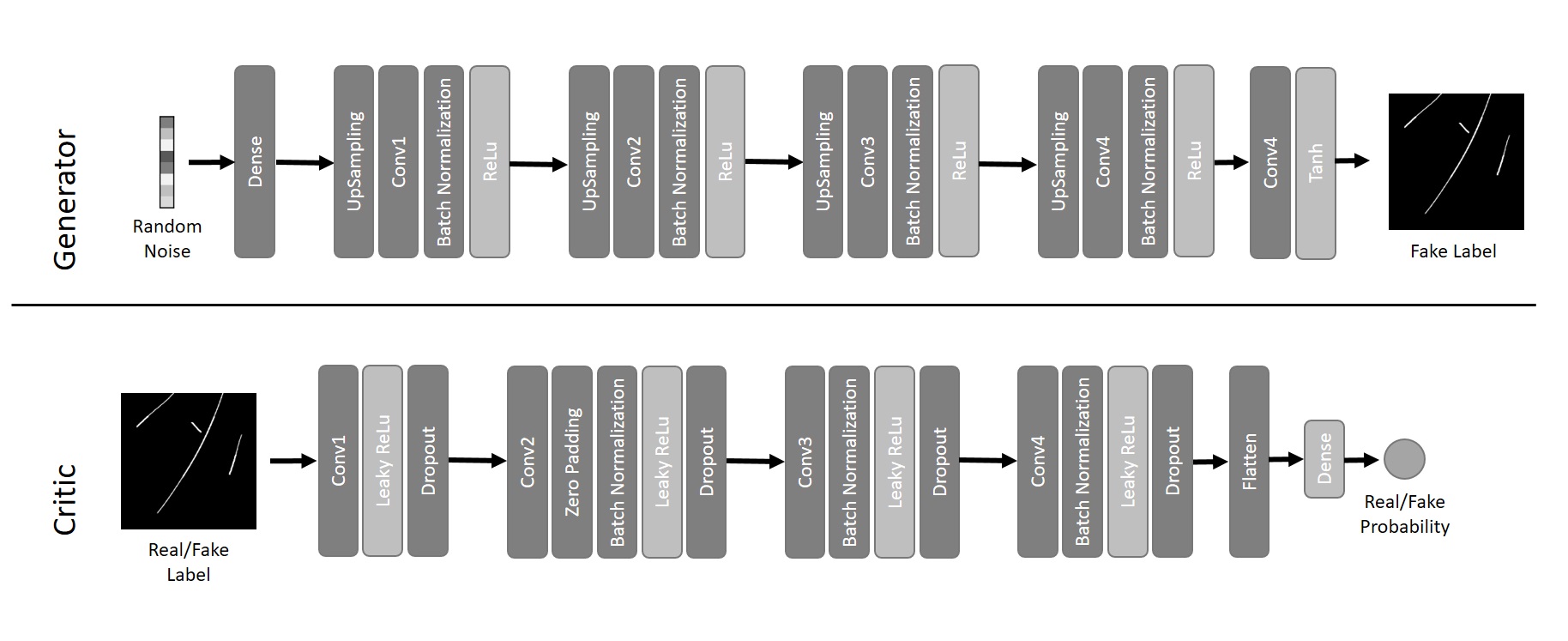}
	\caption{Architecture of both the generator and critic of the used WGAN network.}\label{fig:wgan_architecture}
\end{figure}

The fact that the approach of mathematically modeling label mask structures is coupled to a lot of engineering overhead led to the investigation of a more generic approach, which is described in this section.\\
The basic idea of this method is to use the capability of original GANs to transform random noise vectors into data that looks similar to data of a given training set. While the pix2pix network that was described earlier in this work performs a transformation between different image domains, more traditional GANs are designed to generate completely new data. This property was used for data augmentation tasks in the past, not only in the image domain \cite{bowles2018gan}, but also for audio classification tasks \cite{mmsp2020}.
However, instead of generating new image data of carbon fiber defect images, our approach uses such a rather traditional GAN to create new label segmentation masks, which then can be transformed to defect image data by feeding it to our pix2pix network, as will be described in the next section. The benefit of generating label masks instead of real image data is that, on the one hand, the pixelwise structure of those label masks appears less complex due to its binary nature. On the other hand, the cascade of generating label masks in one step and transforming those label masks further to actual image data in another step makes those image/label pairs available for direct processing: they finally serve as a training set for segmentation networks.\\
To generate those artificial label mask images, we make use of a convolutional GAN that operates on the \textit{Wasserstein-Loss} as introduced by Arjovsky et al. \cite{arjovsky2017wasserstein}.
The network architecture of both the generator and the \textit{critic}, as the discriminator is called in the context of WGANs, is illustrated in Fig.\ref{fig:wgan_architecture}.\\
As a training dataset, we use real label masks. More specifically, the same label masks can be used which were already part of the training pairs of the \textit{pix2pix} network. Details regarding the training configuration can be found in Section \ref{wgan_config}.

\begin{figure*}[t!]
\centering
\includegraphics[width=1.0\textwidth]{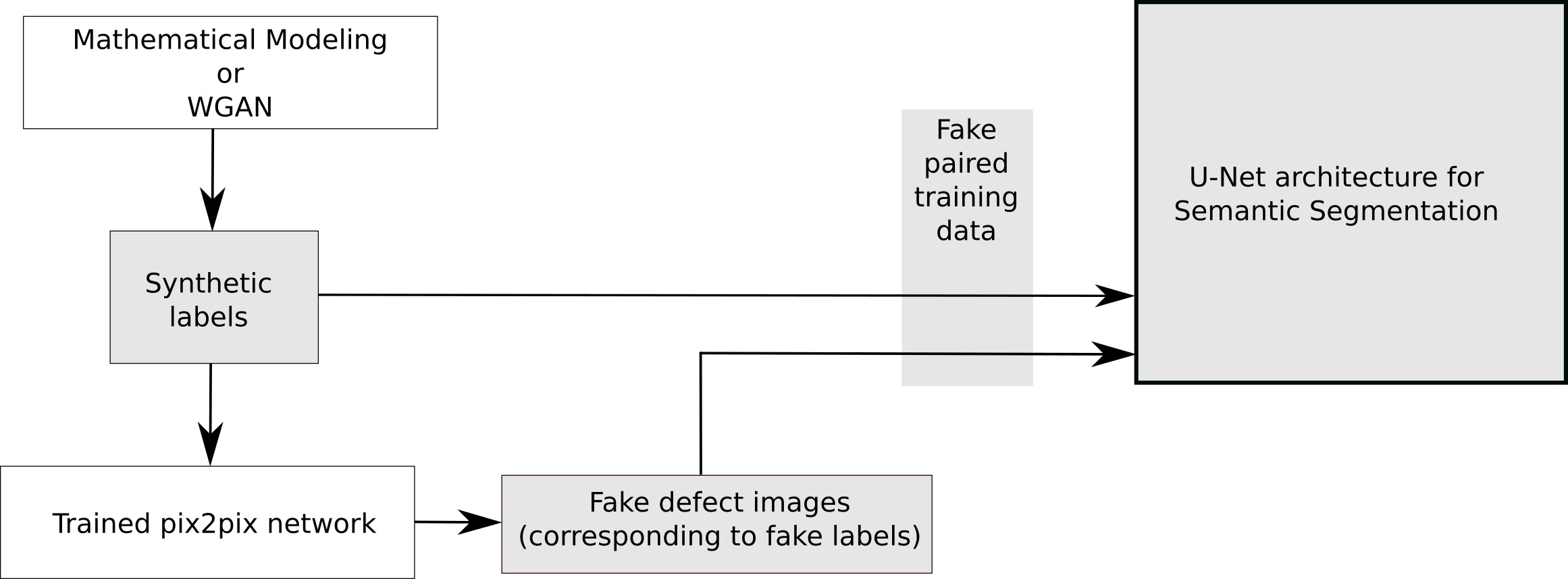}
\caption{The generation and preparation of training data for U-Net using a trained \textit{pix2pix} model and the fake label generator to create fake training pairs. (Step 3) \cite{mertes2020}}\label{fig:step_3}
\end{figure*}

\subsection{Finalizing the Training Data}
In the last step, the generated label data is used to create new corresponding image data. Thus, the label data that was produced by either mathematical modelling or by the WGAN is used as input to the trained \textit{pix2pix} model. The resulting data pairs of label/image data can be used to train further networks for the actual segmentation task. The whole system is shown in Fig. \ref{fig:step_3}. For our experiments, we chose a U-Net architecture to perform this segmentation task. It has to be emphasized that the selection of this specific network was done for the purpose of evaluating and comparing our augmentation approaches, and that the authors don't claim that architecture to be the best choice for the respective task. However, U-Net could achieve promising results in related fields like biomedical image segmentation \cite{ronneberger2015u}.

\begin{figure}
\centering
\includegraphics[width=0.5\textwidth]{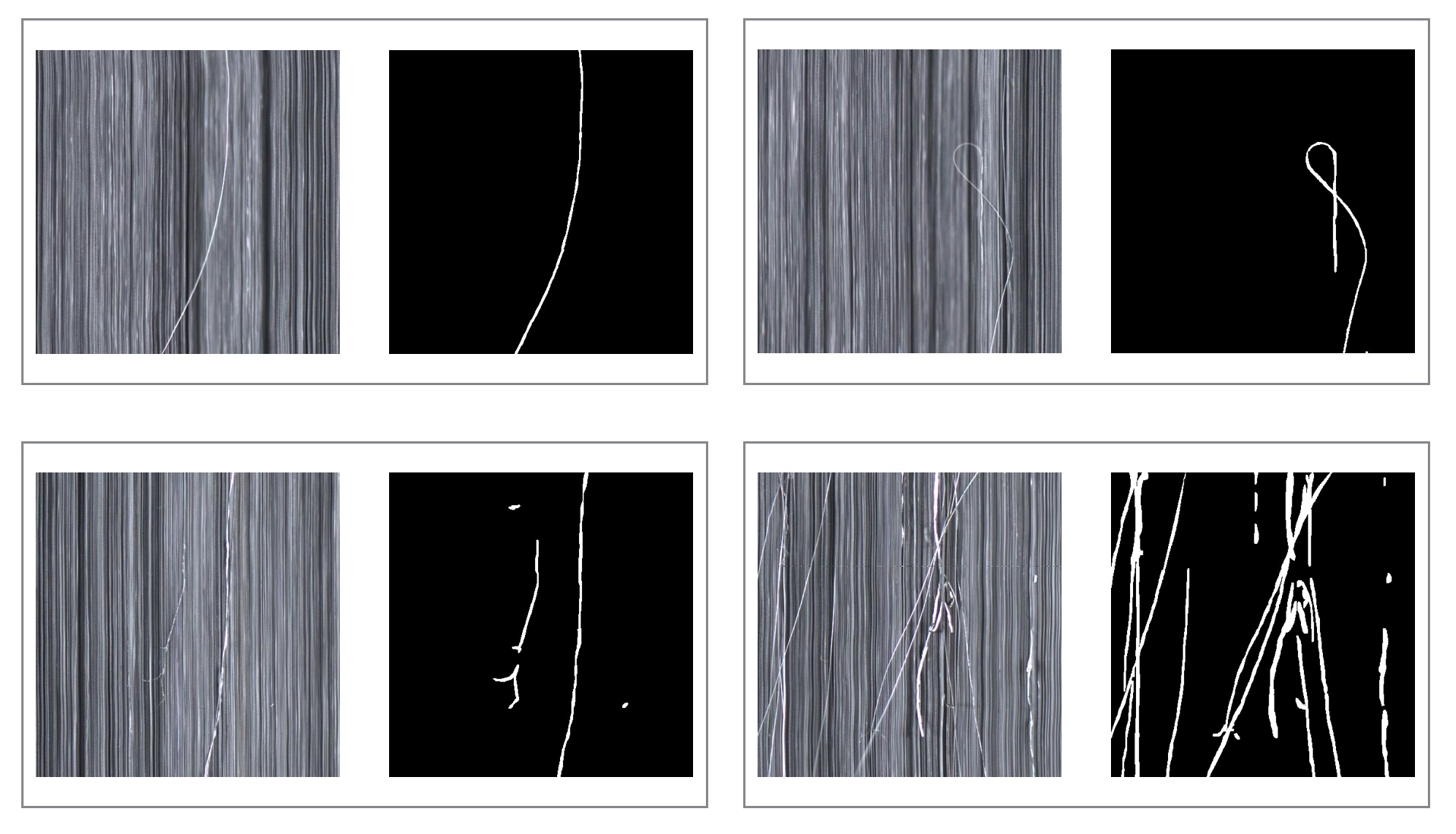}
\caption{Examples of real image data pairs labelled by experts. The misaligned fibers are visible on top of the fiber carpet \cite{mertes2020}.}\label{fig:example_pairs}
\end{figure}

\begin{figure*}[t!]
    \centering
	\begin{subfigure}[t]{0.25\columnwidth}
        \centering
        \includegraphics[width=0.5\columnwidth]{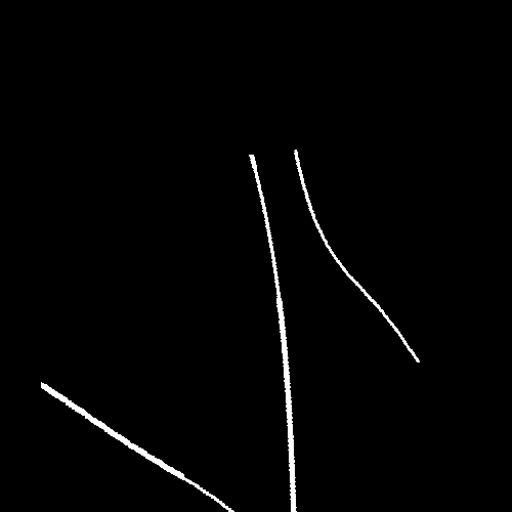}
    \end{subfigure}%
	\begin{subfigure}[t]{0.25\columnwidth}
		\centering
		\includegraphics[width=0.5\columnwidth]{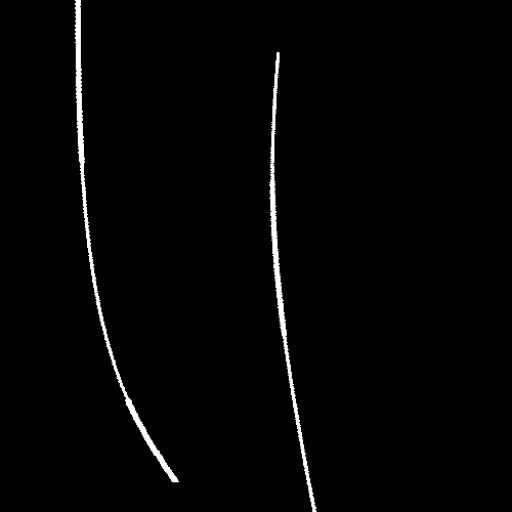}
	\end{subfigure}%
    \begin{subfigure}[t]{0.25\columnwidth}
        \centering
        \includegraphics[width=0.5\columnwidth]{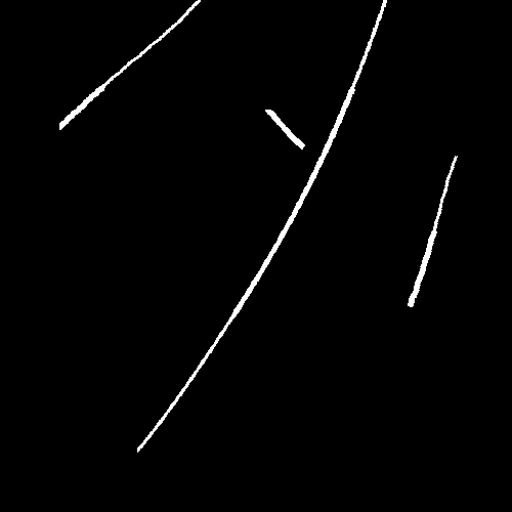}
    \end{subfigure}%
	\begin{subfigure}[t]{0.25\columnwidth}
		\centering
		\includegraphics[width=0.5\columnwidth]{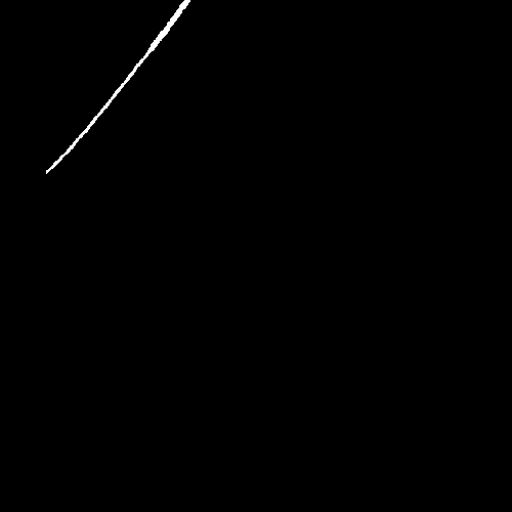}
	\end{subfigure}
	~
    \begin{subfigure}[t]{0.25\columnwidth}
        \centering
        \includegraphics[width=0.5\columnwidth]{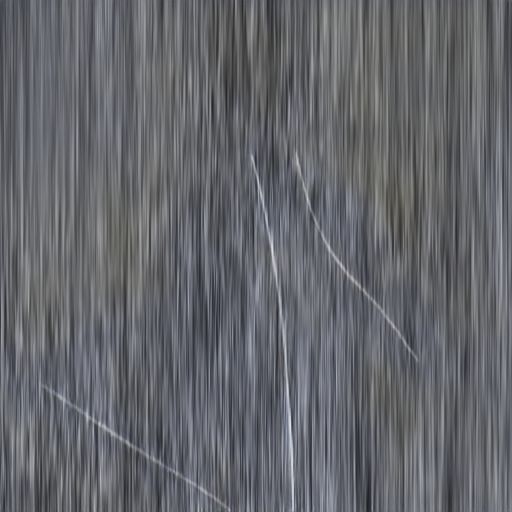}
        \caption{}\label{fig:polygon1binary}
    \end{subfigure}%
	\begin{subfigure}[t]{0.25\columnwidth}
		\centering
		\includegraphics[width=0.5\columnwidth]{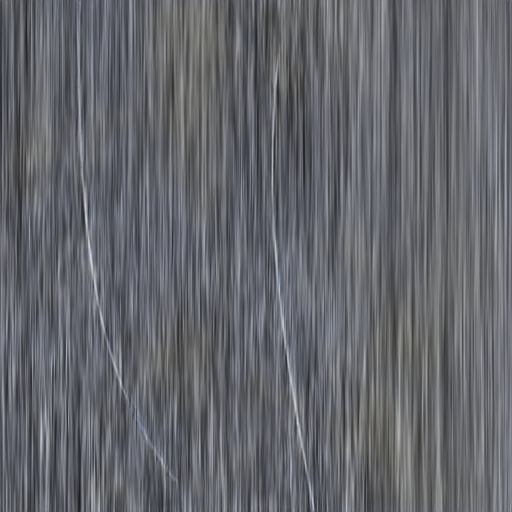}
		\caption{}\label{fig:polygon2binary}
	\end{subfigure}%
    \begin{subfigure}[t]{0.25\columnwidth}
        \centering
        \includegraphics[width=0.5\columnwidth]{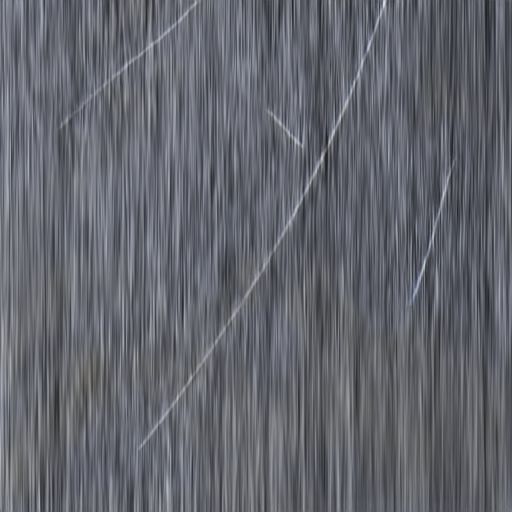}
        \caption{}\label{fig:polygon3binary}
    \end{subfigure}%
	\begin{subfigure}[t]{0.25\columnwidth}
		\centering
		\includegraphics[width=0.5\columnwidth]{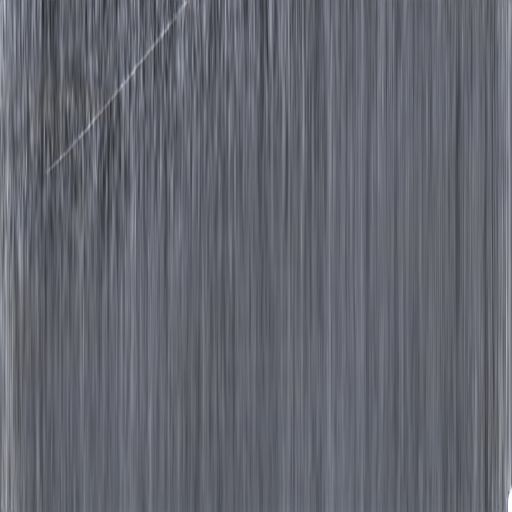}
		\caption{}\label{fig:polygon4binary}
	\end{subfigure}
	\caption{Samples of synthetic labels (top row) and corresponding \textit{pix2pix} outputs (bottom row) imitating misaligned fibers \cite{mertes2020}.}\label{fig:polygons}
\end{figure*}

\begin{figure*}[t!]
    \centering
	\begin{subfigure}[t]{0.25\columnwidth}
        \centering
        \includegraphics[width=0.5\columnwidth]{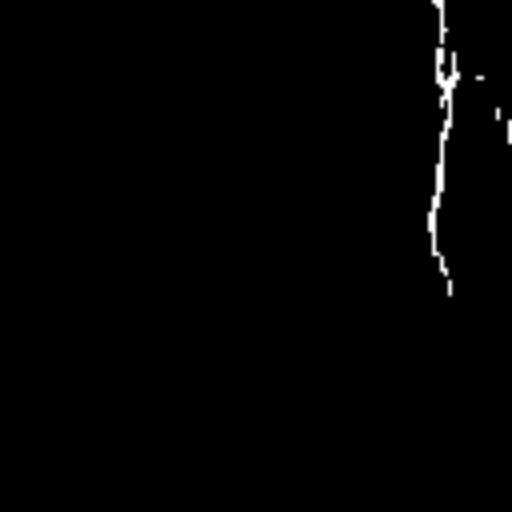}
    \end{subfigure}%
	\begin{subfigure}[t]{0.25\columnwidth}
		\centering
		\includegraphics[width=0.5\columnwidth]{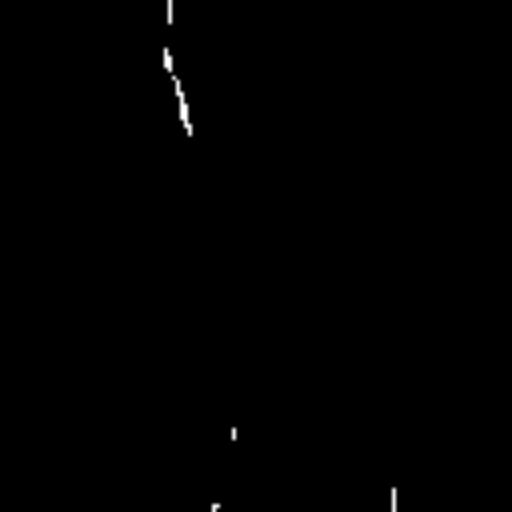}
	\end{subfigure}%
    \begin{subfigure}[t]{0.25\columnwidth}
        \centering
        \includegraphics[width=0.5\columnwidth]{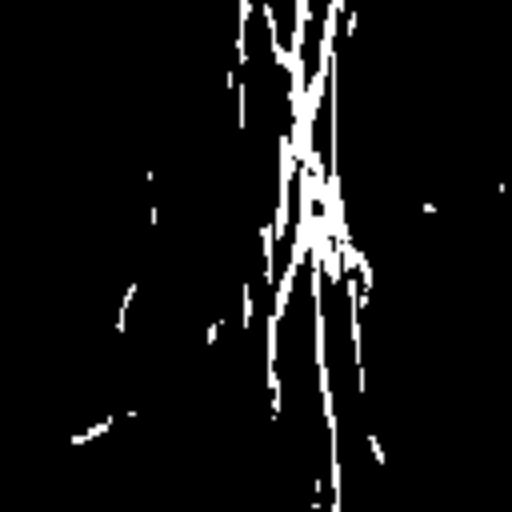}
    \end{subfigure}%
	\begin{subfigure}[t]{0.25\columnwidth}
		\centering
		\includegraphics[width=0.5\columnwidth]{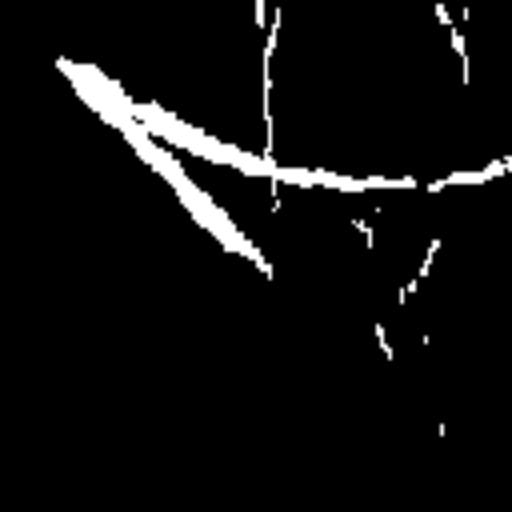}
	\end{subfigure}
	~
    \begin{subfigure}[t]{0.25\columnwidth}
        \centering
        \includegraphics[width=0.5\columnwidth]{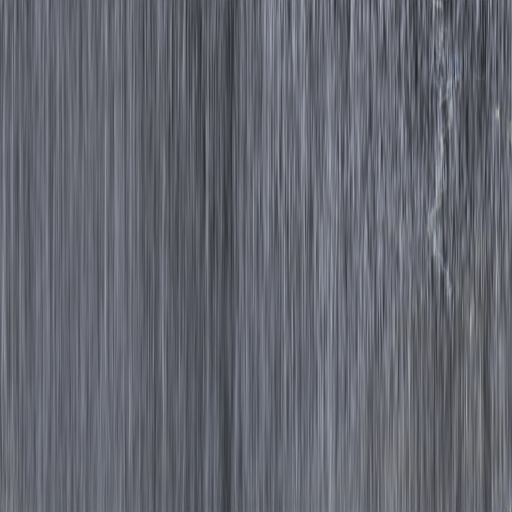}
        \caption{}\label{fig:wganLabels1binary}
    \end{subfigure}%
	\begin{subfigure}[t]{0.25\columnwidth}
		\centering
		\includegraphics[width=0.5\columnwidth]{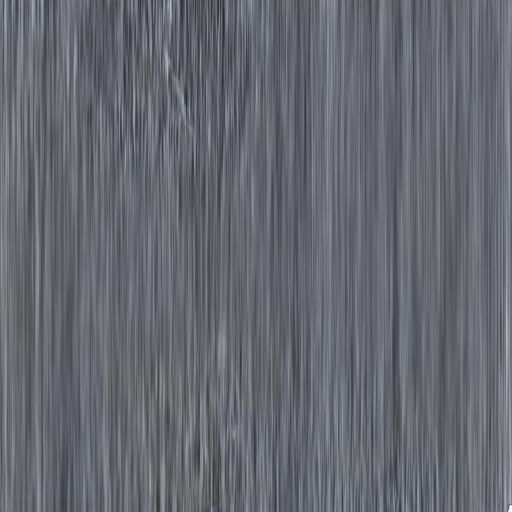}
		\caption{}\label{fig:wganLabels2binary}
	\end{subfigure}%
    \begin{subfigure}[t]{0.25\columnwidth}
        \centering
        \includegraphics[width=0.5\columnwidth]{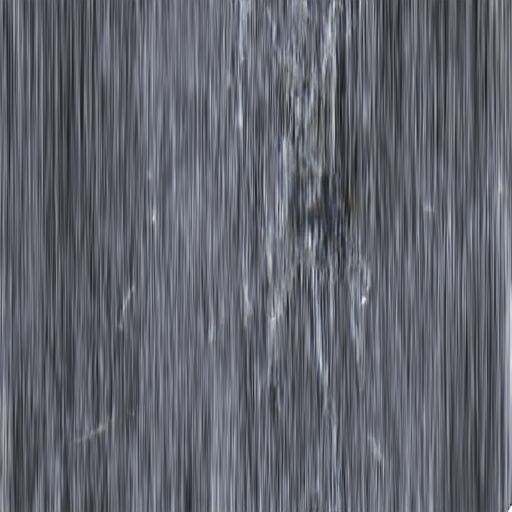}
        \caption{}\label{fig:wganLabels3binary}
    \end{subfigure}%
	\begin{subfigure}[t]{0.25\columnwidth}
		\centering
		\includegraphics[width=0.5\columnwidth]{images/wganPix2pix/train692.jpg}
		\caption{}\label{fig:wganLabels4binary}
	\end{subfigure}
	\caption{Samples of synthetic labels generated by WGAN (top row) and corresponding outputs (bottom row) using the same \textit{pix2pix} model as for Fig. \ref{fig:polygons}.}\label{fig:wganLabels}
\end{figure*}

\section{Experiments and Discussion}\label{sec:experiments}

\subsection{Dataset Specifics}
Our system was evaluated in the context of an industrial application scenario. More precisely, the domain of carbon fiber defect monitoring was chosen for testing and evaluating of the proposed approaches. In images of fiber structures without recognizable defects, the single fibers are aligned in parallel and form a carpet of straight lines. During the production process, mechanical stress caused by spools in the transportation system can lead to damage of the fiber material, which usually can be recognized as misaligned fibers. The shape of those cracked fibers, as well as their position and size vary heavily. Thus, there is no \textit{template} for single defects. Given the different images of defective fiber material, a huge variety of defect structures can be observed. In this specific use case, we aim for the identification of defects on a carbon fiber carpet. To achieve this goal, a U-Net architecture is trained to perform a binary segmentation of pixels that contain defects. Fig. \ref{fig:example_pairs} shows examples of defect images with corresponding binary labels. The environment and the design of the monitoring system that was used to acquire the image data for our experiments has been described in earlier publications ~\cite{geinitz2016linescan,geinitz2016detection,Margraf2017}.

\subsection{Experimental Setup}
For a meaningful evaluation, we ran several experiments to compare the two variants of our approach with conventional data augmentation methods. Thus, parts of our datasets that are described below were augmented with traditional data augmentation techniques. The following \textit{simple} image transformations were applied to those artificially extended datasets: 
\begin{itemize}
    \item Randomised crop of squares of different size (\textit{RandomSizedCrop})
    \item Horizontal and vertical flip
    \item Rotation (for 180 degrees)
    \item Elastic transformation
    \item Grid distortion
\end{itemize}
We arranged the image data in six different sets and performed multiple trainings of a U-Net architecture. Then, we used the resulting models to make predictions on a test set. To ensure comparability, the same test set was used for every training set.  Every training pair for the U-Net architecture consists of a real or fake defect image and a real or fake binary label image. 
The first four datasets are the exact same as used in the publication by Mertes et al. \cite{mertes2020}. We added two additional sets using the WGAN model for generating fake labels as listed below:
\begin{description}
   % \item[Datasets 1 to 4] are based on data taken from the previous experiments presented in \cite{mertes2020}.
    \item[Dataset 1] contains 300 pairs of real defect data and corresponding binary label images. Thus, only original data without DA was taken. 
   \item[Dataset 2] contains the same 300 pairs of defect data and corresponding labels as \textit{dataset 1}. Additionally, an \textit{online} form of data augmentation was applied as described above. For each image some of those aforementioned transformation operations were performed with a predefined probability.
   \item[Dataset 3] contains 3000 pairs of defect data and corresponding labels. 2700 of the 3000 data pairs were generated by applying the \textit{pix2pix} based data augmentation approach on \textit{dataset 1}. For the creation of synthetic label, our mathematical model was applied. Furthermore, the 300 original data pairs from \textit{dataset 1} were taken.
    \item[Dataset 4] contains the same 3000 pairs of defect data and corresponding labels as \textit{dataset 3}. Additionally, the same conventional stochastic data augmentation as for \textit{dataset 2} was applied, i.e. each image was transformed with a predefined probability during training. Thus, \textit{dataset 4} combines common data augmentation with our approach. 
    
    \item[Dataset 5] contains 3000 pairs of defect data and corresponding labels. In this dataset, 2700 of the 3000 pairs were generated by the \textit{pix2pix} network. This time, however, the input data for image-to-image-translation were not generated from a trigonometic function, but by training a \textit{WGAN} model on image pairs of real sample data. The resulting model was then used to create new binary labels. The remaining 300 image pairs were taken from \textit{dataset 1} as performed in the previous datasets.
    
    \item[Dataset 6] contains the same 3000 pairs of defect data and corresponding labels as \textit{dataset 5}. Hereby, though, we altered the training of the U-Net by adding \textit{online, stochastic} data augmentation as for \textit{dataset 4} and \textit{dataset 2}. In this dataset, 2700 of the 3000 pairs were generated by the \textit{pix2xpix} network. This was performed to examine how \textit{regular} data augmentation will change the result on top of WGAN based label generation and image-to-image translation.
\end{description}

Each of the datasets was used to train a separate U-Net model for semantic segmentation. For testing and evaluation, a single, distinct dataset was used, containing real defect data and annotations that are provided by domain experts.

\subsection{Pix2Pix Configuration}
The configuration of the \textit{pix2pix} model is given in Tab. \ref{tab:pix2pixconfig}. We stopped the training after 3200 epochs, as we could not observe any further improvement of the generated images by that time. Fig. \ref{fig:polygons} shows a selection of pairs of labels and images generated through application of the \textit{pix2pix} model, where the labels where created by mathematical modeling. Fig. \ref{fig:wganLabels} shows label/image pairs where the labels were generated by our WGAN approach.

\begin{table}[ht]
\centering
\caption{Pix2Pix Configuration \cite{mertes2020}}
 \begin{tabular}{|c | c |}
 \hline
 Parameter & Value \\ [0.5ex] 
 \hline\hline
 Learning rate & 0.0005\\
 Batch Size & 1\\
 Epochs & 3200\\
 Loss Function & Mean Squared/Absolute Error\\
 \hline
\end{tabular}\label{tab:pix2pixconfig}
\end{table}

\subsection{U-Net Configuration}
The U-Net architecture was trained individually for every dataset. As described above, dataset 2, 4 and 6 were augmented with traditional DA, i.e. conventional image transforms.

All of those conventional data augmentation methods are based on the library published by \cite{rizki2018}. We used the same methods that were already described in \cite{mertes2020}. A stochastic component was added to the image transformations, i.e. all operations were performed with a given probability. \\
The randomized crop was given the probability $p = 0.25$ and a window size interval of $[400,512]$ pixels. Furthermore, the probabilities for flipping, rotation, elastic transform and grid distortion were set to $p = 0.5$. For these three methods, only one operation, i.e. either elastic transform or grid distortion was allowed (\textit{OneOf}). The degree of rotation was set to exactly $180$, as the structures of carbon fibres used in our experiments always have a vertical alignment. \textit{Elastic Transform} was performed with the parameters $\alpha = 10$, $\sigma = 10$, $alpha\_affine = 512 \cdot 0.05$ and $border\_mode = 4$. \textit{Grid Distortion} was given the parameters $num\_steps = 2$ and $distort\_limit = 0.4$. In the respective datasets, the operations were applied online using the given parameters on every original input image. 
The U-Net model itself was slightly adapted from \cite{Yakubovskiy2019} to fit the dataset. The default size of the training images was $512x512$,
yet the default U-Net setting only accepts images of size $28x28$. A ResNet-18 model is used as encoder by the U-Net. The architecture was adapted to fit the input size before applying the model. The training configuration of the U-Net is shown in Tab. \ref{tab:unetconfig}.
\begin{table}[ht]
\centering
\caption{U-Net Configuration \cite{mertes2020}}
 \begin{tabular}{|c | c |}
 \hline
 Parameter & Value \\ [0.5ex] 
 \hline\hline
 Learning rate & 0.0001\\
 Batch Size &  10\\
 Epochs & 200\\
 Loss Function & Binary Cross Entropy / Dice Loss\\
 \hline
\end{tabular}\label{tab:unetconfig}
\end{table}

\subsection{WGAN Configuration}\label{wgan_config}
The configuration for the WGAN training is shown in Table \ref{tab:wganconfigtable}. As the WGAN produces non-binary image data as output, we applied a binarization stage to the final output images in order to get binary label masks. 
\begin{table}[ht]
\centering
\caption{WGAN Configuration}
 \begin{tabular}{|c | c |}
 \hline
 Parameter & Value \\ [0.5ex] 
 \hline\hline
 Learning rate & 0.00005\\
 Batch Size & 64\\
 Epochs & 100,000\\
 \hline
\end{tabular}\label{tab:wganconfigtable}
\end{table}

\begin{table*}[t]
\centering
\caption{U-Net results from test runs on the datasets 1 through 6 for batch size 5, based on \cite{mertes2020}.}
 \begin{tabular}{|c || c c c c c c c ||}
 \hline
{} & PPV & TPR & IoU & ACC & MCC & F1 & F2 \\ [0.5ex] 
 \hline\hline
Dataset 1 & 0.539169 & 0.586753 & 0.390778 & 0.985035 & 0.55487 & 0.561956 & 0.576576 \\
Dataset 2 & 0.772803 & 0.718101 & 0.592925 & 0.991935 & 0.740872 & 0.744448 & 0.728413 \\
Dataset 3 & 0.745926 & 0.721067 & 0.578888 & 0.991419 & 0.729034 & 0.733286 & 0.725905 \\
Dataset 4 & 0.756767 & 0.705387 & 0.57502 & 0.991471 & 0.72631 & 0.730175 & 0.715098 \\
Dataset 5 & 0.602418 & 0.585941 & 0.422541 & 0.990628 & 0.594065 & 0.543877 & 0.589383 \\
Dataset 6 & 0.281570 & 0.433361 & 0.205801 & 0.980426 & 0.339847 & 0.341352 & 0.354881 \\
 \hline
\end{tabular}\label{tab:datasets}
\end{table*}

\section{Evaluation}\label{sec:discussion}

The metrics $accuracy$, $MCC$ and $F_\beta-Score$ are less dependable for an objective evaluation of the classification model in our specific task. The proportion between background pixels (i.e. the non-defect pixels) and foreground pixels (i.e. the defect pixels) per image is thoroughly unbalanced, as the defects mostly consist of single fibers and therefore take much less space in the images. While $accuracy$ returns the proportion of true results among all data points examined, $MCC$ and $F_\beta-Score$ aim to balance out true and false positives and negatives of the binary classification result.
In contrast, the \textit{Jaccard index} or \textit{Intersection over Union (IoU)} is used to measure the similarity of two sets, i. e. the similarity of the ground truth and the prediction. This property makes the $IoU$ the most appropriate for the task at hand.
%It is defined as the area of overlap over the area of union or formally: 
%\begin{equation}
%    IoU = \frac{|A \cap B|}{|A \cup B|}
%\end{equation}
Thus, we focus on the $IoU$ metric for our experiments in order to allow an objective and problem related evaluation methodology. However, all relevant statisticals scores are reported in Tab. \ref{tab:datasets} for the sake of completeness.

\subsection{Discussion of Results}
\begin{figure}
	\includegraphics[width=1.0\columnwidth]{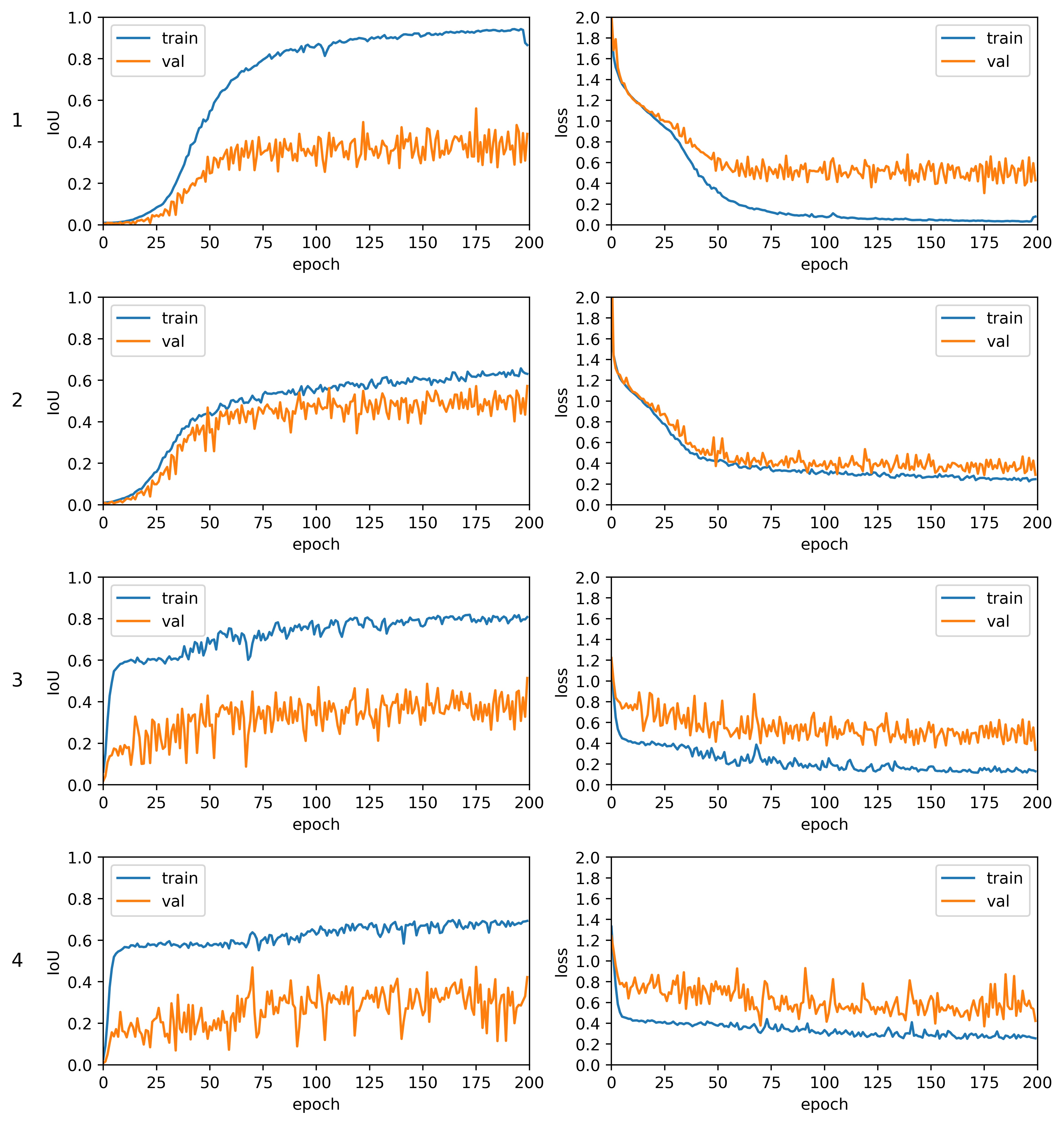}
	\caption{IoU Score and Loss during U-Net training for dataset 1 through 4 from top to bottom row \cite{mertes2020}.}\label{fig:training_loss}
\end{figure}

For all of the six datasets the training was aborted after $200$ epochs since it was clearly observable that the models had converged. As Fig. \ref{fig:training_loss} suggests, both training accuracy and loss converge from epoch 100 onwards for the first four datasets. Training on dataset 1 was stopped at a loss rate of $\sim 0.7$, while for both datasets 3 and 4 the training ended at a loss rate of $\sim 0.4$. For dataset 2, model training reached an IoU of $\sim 0.6$ and $\sim 0.5$ for validation when the process was aborted. At the same time, the training loss ended at $\sim 0.4$ and reached a value of $\sim 0.2$ for validation. Furthermore, training on dataset 1 reached an IoU score of $\sim 0.7$ while dataset 3 and 4 achieved an IoU value of $\sim 0.8$ after 200 epochs. 

\begin{figure}
	\begin{subfigure}[t]{0.5\columnwidth}
		\centering
		\includegraphics[width=1.0\columnwidth]{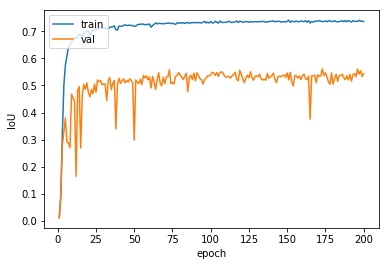}
		%\caption{}
	\end{subfigure}
	\begin{subfigure}[t]{0.5\columnwidth}
		\centering
		\includegraphics[width=1.0\columnwidth]{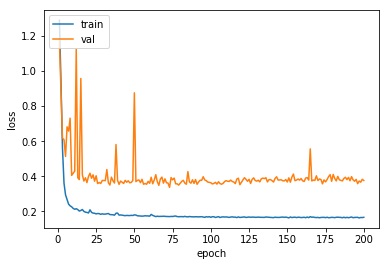}
		%\caption{153}\label{fig:raw153}
	\end{subfigure}
	~
	\begin{subfigure}[t]{0.5\columnwidth}
		\centering
		\includegraphics[width=1.0\columnwidth]{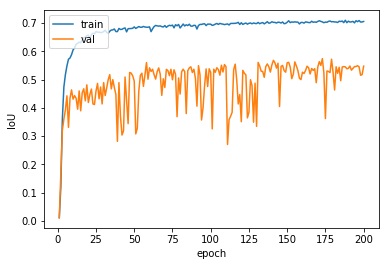}
		%\caption{153}\label{fig:raw153}
	\end{subfigure}
	\begin{subfigure}[t]{0.5\columnwidth}
		\centering
		\includegraphics[width=1.0\columnwidth]{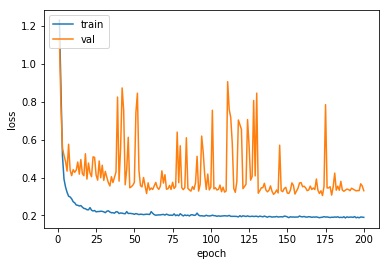}
		%\caption{153}\label{fig:raw153}
	\end{subfigure}
	\caption{IoU Score and Loss during U-Net training for dataset 5 and 6 from top to bottom row.}\label{fig:training_loss_2}
\end{figure}

Also, Fig. \ref{fig:training_loss_2} shows that training on dataset 5 and 6 reached an $IoU$ score of $\sim 0.7$ and $\sim 0.5$ for validation after 200 epochs. In both cases, the loss converged at $\sim 0.2$ for training and $\sim 0.4$ for validation. All training results for the four datasets are shown in Tab. \ref{tab:datasets}.

% ============
% Test results
% ============
The trained models were alltogether evaluated on the test dataset. The model trained on dataset 1 reached an $accuracy$ of $0.985$ and $IoU$ of $0.391$ on the test set, while the model trained with dataset 2 reached an $accuracy$ of $0.992$ and an $IoU$ of $0.593$. Furthermore, the $IoU$ for the model based on dataset 3 reached an IoU of $0.579$ and an $accuracy$ of $0.991$, while training with dataset 4 achieved a value of $0.575$ for the $IoU$ and $0.991$ for the $accuracy$. In addition, dataset 5 resulted in an IoU of $0.423$ and $accuracy$ of $0.991$, while training on dataset 6 let the $IoU$ drop to $0.206$ and $accuracy$ decrease to $0.980$.
All metrics for the test runs were acquired from prediction on 25 randomly selected sample images as presented in Tab. \ref{tab:datasets}.

\begin{figure*}[t!]
    \centering
    \begin{subfigure}[t]{0.25\columnwidth}
        \centering
        \includegraphics[width=0.8\columnwidth]{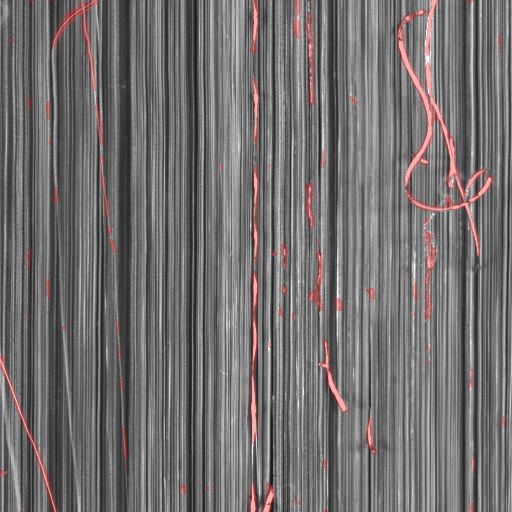}
        %\caption{69}\label{fig:raw69}
    \end{subfigure}%
	\begin{subfigure}[t]{0.25\columnwidth}
		\centering
		\includegraphics[width=0.8\columnwidth]{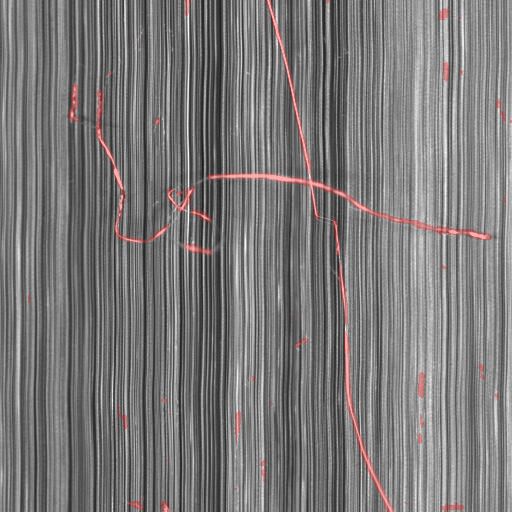}
		%\caption{21}\label{fig:raw21}
	\end{subfigure}%
    \begin{subfigure}[t]{0.25\columnwidth}
        \centering
        \includegraphics[width=0.8\columnwidth]{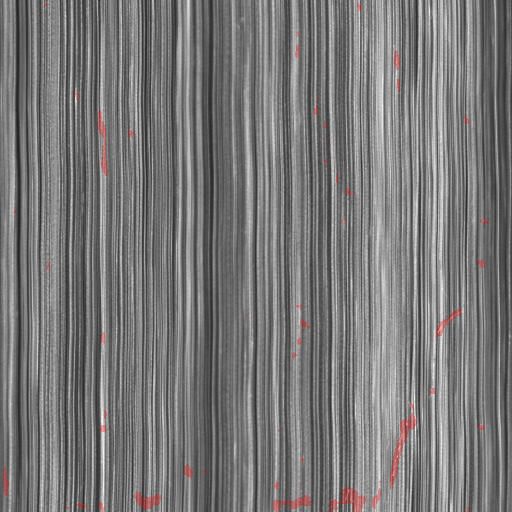}
        %\caption{111}\label{fig:raw111}
    \end{subfigure}%
	\begin{subfigure}[t]{0.25\columnwidth}
		\centering
		\includegraphics[width=0.8\columnwidth]{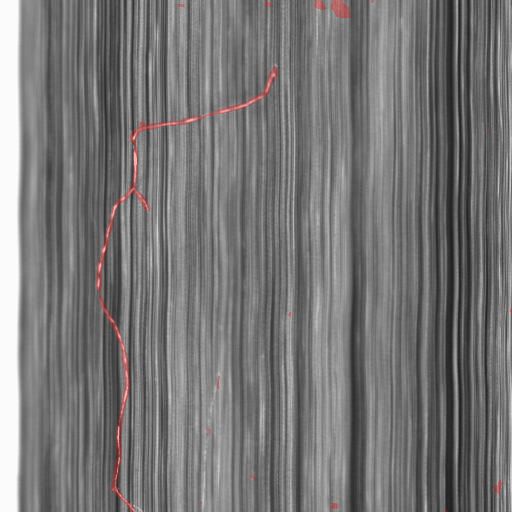}
		%\caption{153}\label{fig:raw153}
	\end{subfigure}
	~
    \begin{subfigure}[t]{0.25\columnwidth}
        \centering
        \includegraphics[width=0.8\columnwidth]{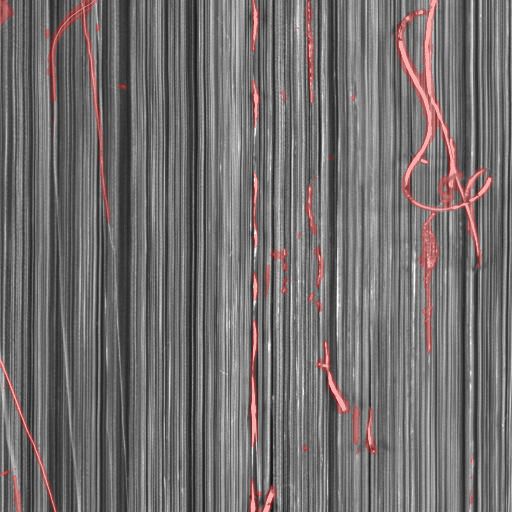}
        %\caption{69}\label{fig:aug69}
    \end{subfigure}%
	\begin{subfigure}[t]{0.25\columnwidth}
		\centering
		\includegraphics[width=0.8\columnwidth]{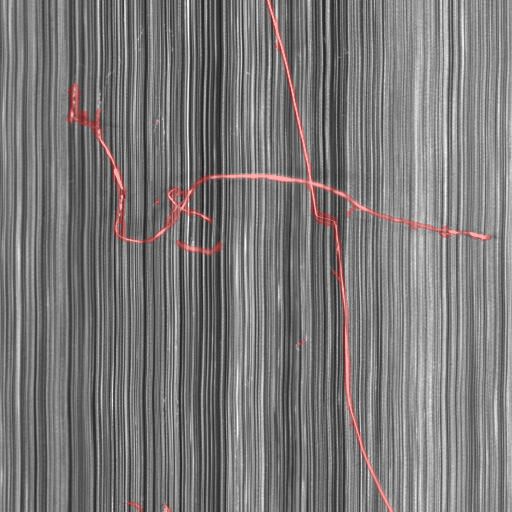}
		%\caption{21}\label{fig:aug21}
	\end{subfigure}%
    \begin{subfigure}[t]{0.25\columnwidth}
        \centering
        \includegraphics[width=0.8\columnwidth]{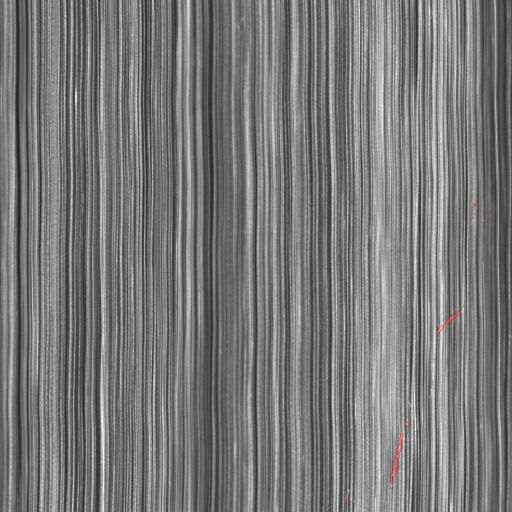}
        %\caption{111}\label{fig:aug111}
    \end{subfigure}%
	\begin{subfigure}[t]{0.25\columnwidth}
		\centering
		\includegraphics[width=0.8\columnwidth]{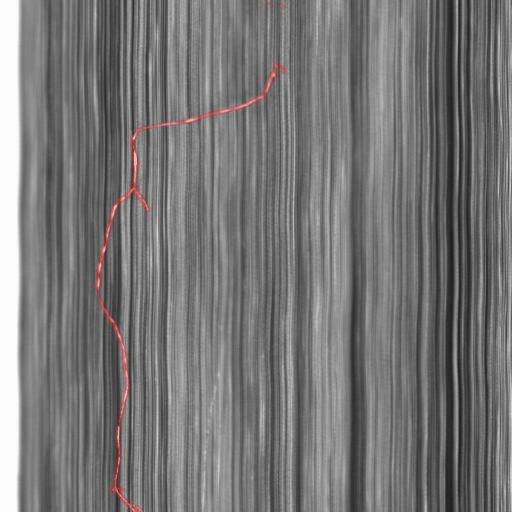}
		%\caption{153}\label{fig:aug153}
	\end{subfigure}
	~
    \begin{subfigure}[t]{0.25\columnwidth}
        \centering
        \includegraphics[width=0.8\columnwidth]{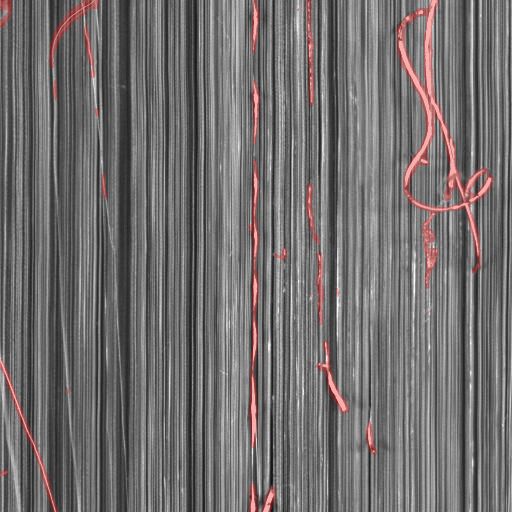}
        %\caption{69}\label{fig:wganAug69}
    \end{subfigure}%
	\begin{subfigure}[t]{0.25\columnwidth}
		\centering
		\includegraphics[width=0.8\columnwidth]{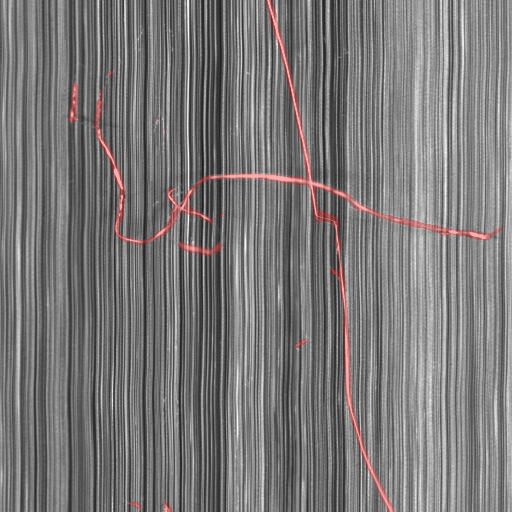}
		%\caption{21}\label{fig:wganAug21}
	\end{subfigure}%
    \begin{subfigure}[t]{0.25\columnwidth}
        \centering
        \includegraphics[width=0.8\columnwidth]{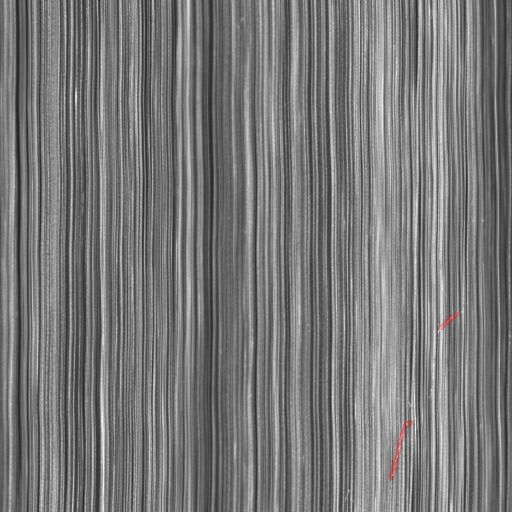}
        %\caption{111}\label{fig:wganAug111}
    \end{subfigure}%
	\begin{subfigure}[t]{0.25\columnwidth}
		\centering
		\includegraphics[width=0.8\columnwidth]{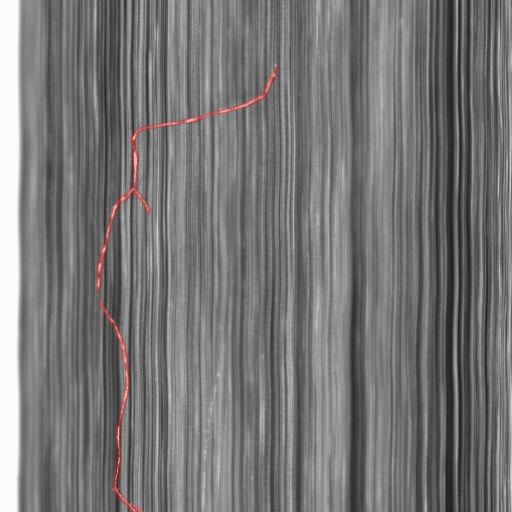}
		%\caption{153}\label{fig:wganAug153}
	\end{subfigure}
	~
    \begin{subfigure}[t]{0.25\columnwidth}
        \centering
        \includegraphics[width=0.8\columnwidth]{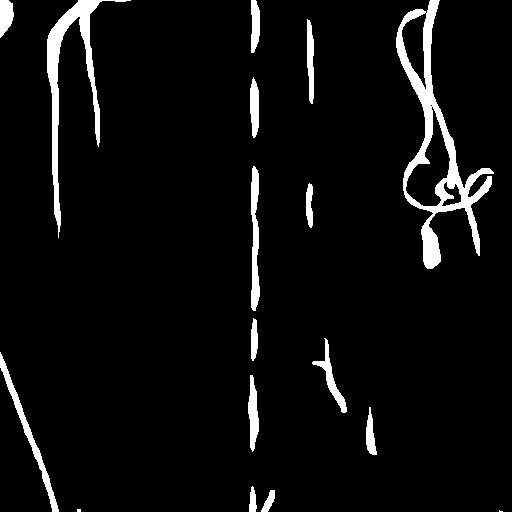}
        \caption{}\label{fig:mask69}%69
    \end{subfigure}%
	\begin{subfigure}[t]{0.25\columnwidth}
		\centering
		\includegraphics[width=0.8\columnwidth]{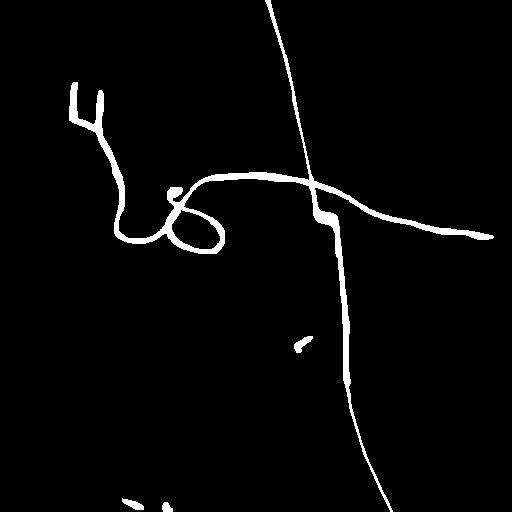}
		\caption{}\label{fig:mask21}%21
	\end{subfigure}%
    \begin{subfigure}[t]{0.25\columnwidth}
        \centering
        \includegraphics[width=0.8\columnwidth]{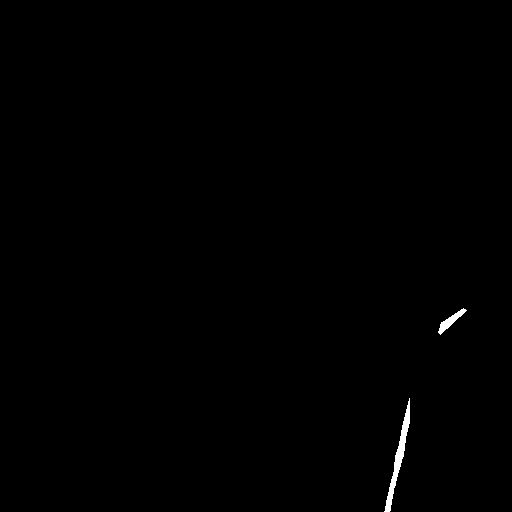}
        \caption{}\label{fig:mask111}%111
    \end{subfigure}%
	\begin{subfigure}[t]{0.25\columnwidth}
		\centering
		\includegraphics[width=0.8\columnwidth]{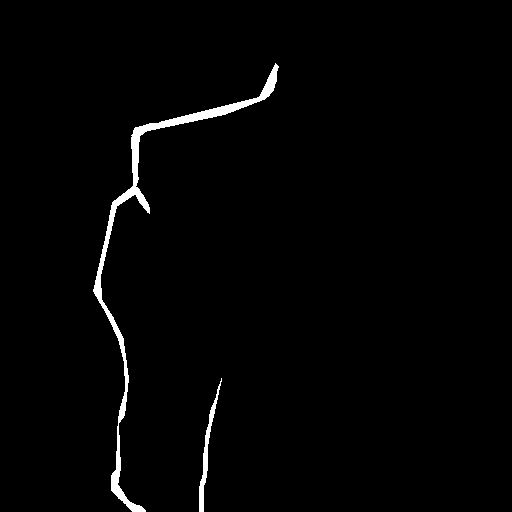}
		\caption{}\label{fig:mask153}%153
	\end{subfigure}%
	\caption{Real carbon fiber defects from the test set with red overlay from U-Net segmentation for \textit{dataset 1} (top row), \textit{dataset 3} (second row), \textit{dataset 5} (third row) and the \textit{ground truth} (bottom row) \cite{mertes2020}}\label{fig:raw_results}
\end{figure*}

Fig. \ref{fig:raw_results} shows a sample selection of defect images taken from the test set with red overlays representing the ROIs predicted by the U-Net model. It should be noted that training without any DA leads to more false positives which reminds of noise in the overlays as can be seen in Fig. \ref{fig:mask21} and \ref{fig:mask111}.

For \textit{dataset 1}, the loss rate drops heavily for 50 epochs before converging at a value of around $0.5$ for the test set and just over $0.0$ for the training set, as can be seen in Fig. \ref{fig:training_loss}. The $IoU$ value increases for $50$ epochs before slowing down and converging to an $IoU$ of around $0.85$ for the training set and around $0.4$ for the test set after 125 epochs.

For \textit{dataset 2} the loss rate drops considerably during the first 5 epochs, then decreases slowly but steady until it converges around $0.2$ after 125 epochs during training. The loss on the test set shows a similar behaviour, except it converges around a value of $0.4$. Again, the $IoU$ value increases clearly within less than 5 epochs during the training process, then only slightly rises before converging around $0.6$ after epoch 125. For the test set, the $IoU$ value increases constantly between epoch $0$ and $75$, then converges around $0.4$ as illustrated in the second row of Fig. \ref{fig:training_loss}.

As can be seen in Fig. \ref{fig:training_loss_2}, the loss rate of dataset 5 drops heavily for 25 epochs, then quickly converges around $0.2$ for the training set. The loss curve behaves the same way, although it oscillates more. The documented loss on dataset 6 shows the same characteristics, however its loss curve oscillates even more on the test set.

\subsection{Assessment of Results and Insights}
As can be seen, the U-Net model trained on dataset 3 significantly outperformed the model trained on dataset 1. This shows that our approach of mathematical defect modeling in combination with a \textit{pix2pix} architecture could substantially improve the quality and diversity of the raw training set. When comparing the results of dataset 2 and dataset 3, it becomes apparent that the proposed approach is slightly worse, however not significantly diverts from conventional data augmentation techniques as applied on dataset 2. The difference comprises within less then $0.02$ for the \textit{IoU}. 

The combination of generating synthetic data using that first approach with subsequent conventional data augmentation as for dataset 4 did not lead to any improvement. The model trained on dataset 4 outperforms dataset 1, but leads to a slighty lower \textit{IoU}, \textit{accuracy} and \textit{MCC} than dataset 2 and 3. However, the degradation ranges within less then $0.02$ for the \textit{IoU} and is therefore not significant under the given circumstances.
% =========
% Dataset 5
% =========
The last two rows of Tab.  \ref{tab:datasets} show the results from the experiments that apply image-to-image translation on labels generated with the WGAN model. As can be seen, dataset 5 slightly outperforms dataset 1 since it results in a higher \textit{IoU}, \textit{accuracy} and \textit{MCC}. However, it clearly proves inferior to datasets 2 through 4 which means regular data augmentation as well as the problem-tailored label generator clearly performs better in the given scenario.
% =========
% Dataset 6
% =========
For the sake of completeness and transparency, we also did an experiment in which we extended the data of \textit{dataset 5} with conventional data augmentation, resulting in \textit{dataset 6}. The outcome, though, indicates a clear deterioration all along the line. Every metric appears lower then for any other dataset (i. e. dataset 1 - 5 perform better). \textit{Precision}, \textit{IoU}, \textit{MCC}, $F_{1}-Score$ and $F_{2}-Score$ drop heavily - to values of $0.28$, $0.21$, $0.34$, $0.34$ and $0.35$. The only metrics, that show a comparably `light' degradation are \textit{accuracy} (to 0.98) and \textit{recall} (to 0.43) which is due to the fact that they weigh out background and foreground pixels. Since only $\sim 1 \%$ of the pixels are associated with the region of interest, the $accuracy$ and $recall$ cannot provide high significance.

The experiment based on dataset 6 obviously suggests that online data augmentation does not add any value to WGAN generated labels and image-to-image translation. The numbers suggest that it even worsens the performance of the semantic segmentation model by a magnitude.

% ===================
% Critical Reflection
% ===================
As the results from Tab. \ref{tab:datasets} and the samples depicted in Fig. \ref{fig:raw_results} suggest, the pairs of synthetic images and labels of carbon fiber defects were successfully used to replace traditional data augmentation for semantic segmentation network training. With an $IoU$ of $0.579$, the first variant of our approach, i.e. the mathematical modeling of defects, performs comparably to U-Net training with regular data augmentation. Since the absolute difference between dataset 2 and 3 results in a value of $0.01$, it appears negligible. The results show that synthesized training data helps to improve the detection quality of a U-Net segmentation model to a great extent. Moreover, the augmented dataset could be created based on few samples of only 300 images with an image size of $512x512$ pixels in which the \textit{ROI} on average only covers ~1 \% of an image frame.

\begin{figure*}[t!]
    \centering
	\begin{subfigure}[t]{0.3\columnwidth}
        \centering
        \includegraphics[width=0.8\columnwidth]{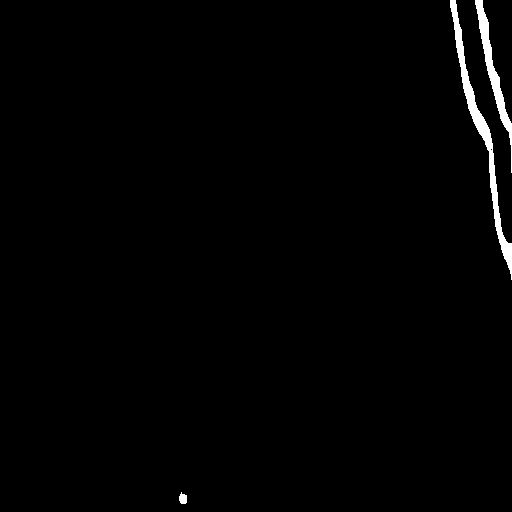}
        %\caption{Ground Truh}
    \end{subfigure}%
	\begin{subfigure}[t]{0.3\columnwidth}
		\centering
		\includegraphics[width=0.8\columnwidth]{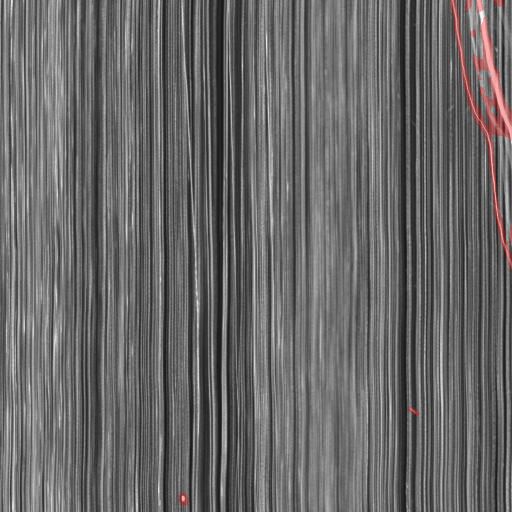}
		%\caption{No DA}
	\end{subfigure}%
    \begin{subfigure}[t]{0.3\columnwidth}
        \centering
        \includegraphics[width=0.8\columnwidth]{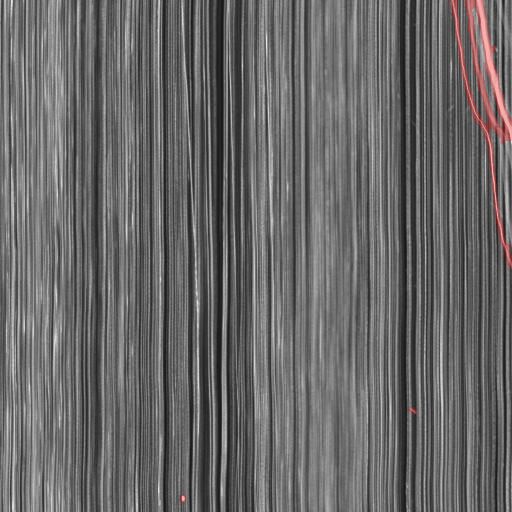}
        %\caption{With DA}
    \end{subfigure}
	~
    \begin{subfigure}[t]{0.3\columnwidth}
        \centering
        \includegraphics[width=0.8\columnwidth]{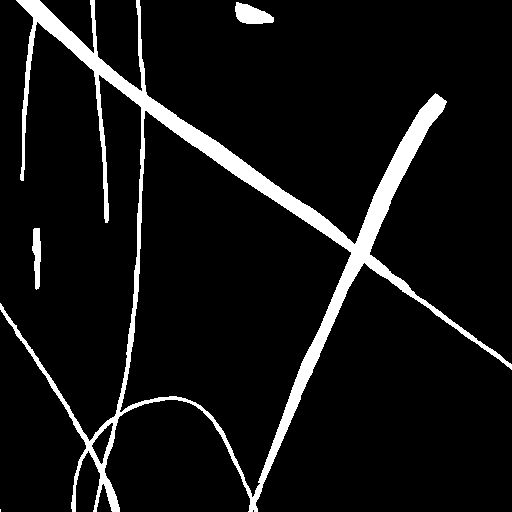}
        \caption{Ground Truth}
    \end{subfigure}%
	\begin{subfigure}[t]{0.3\columnwidth}
		\centering
		\includegraphics[width=0.8\columnwidth]{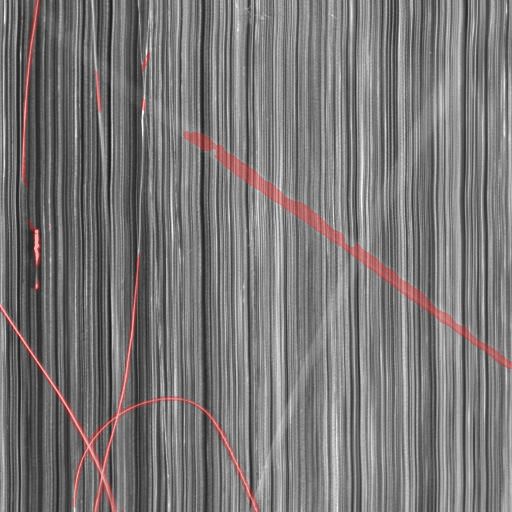}
		\caption{No DA}
	\end{subfigure}%
    \begin{subfigure}[t]{0.3\columnwidth}
        \centering
        \includegraphics[width=0.8\columnwidth]{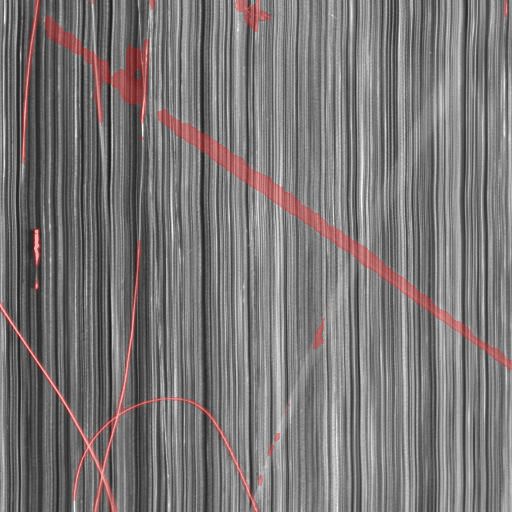}
        \caption{With DA}
    \end{subfigure}
	\caption{Comparison of noticable side effects in the classification of filaments based on the WGAN based approach with and without data augmentation and the corresponding ground truth}\label{fig:wganDifferences}
\end{figure*}

\subsection{Visual Comparison and Critical Reflection}
Since size of a ROI only take up a small portion of each image, even little variance in the detection results heavily affects the evaluation. This is the case if the classification of single pixels differs slightly from the ground truth. Adding a small margin to each filament e. g. largely affects metrics such as $IoU$ or $MCC$ and blurs the results of a proper analysis in this particular case. As can be seen in Fig. \ref{fig:raw_results}, the visual comparison of test results suggests that WGAN+Pix2Pix based semantic segmentation tends to return better segmentation than training a U-Net model without any data augmentation.

Actually, the ground truth sometimes reveals some room for misinterpretation by showing annotations that results in contradicatory objective functions: as can be seen in Fig. \ref{fig:wganDifferences}, the ground truth differs substantially from all segmentation results; to the user this may not be obvious and visually the results range at a very similiar level, but little variations at this level cause the metrics to fluctuate heavily. Although this is intended to some extent, it must be considered closely upon interpretation. All in all, we consider this approach as promising and stress that future work should test it on other similiar or very different applications.

Moreover, the authors suggest to keep in mind that this study has only applied the workflow based on WGAN, Pix2Pix and trigonometric functions on images of a carbon fiber surface with its very special types of defects and anomalies. At this point, it cannot be ruled out that the workflow functions differently under other circumstances that are considerably distinct to the present use case. In fact, the authors greatly encourage the research community to test this approach on other datasets and settings.

We consider both our \textit{pix2pix} based image generation approaches a more realistic and application-oriented form of DA. The experiments were conducted with and without traditional DA in order to evaluate the effectiveness and expandability of our approach. 
% ===========
% Overfitting 
% ===========
Excessive use of traditional DA might superimpose the `real' data within the training set due to its low level form of manipulation, reproduction and reuse which raises the risk of overfitting during model training. GAN based data generation is also less prone to repetitive patterns since it tries to project the variation found in the original data to the synthetic data.

\subsection{Summary}
In summary, the approach as proposed in this article reveals a potential alternative to traditional, simple data augmentation. The generated data forms representation of images that are significantly different from the sample data but still resemble the training distribution. Furthermore, the suggested algorithms support training deep learning models for semantic segmentation with small sample datasets. This also applies if only few annotations are available.

In this article, two competing concepts were suggested and evaluated on industry data. While the first approach based on a trigonometric function represents a very problem-specific, hand-crafted solution, the second concept entirely uses adversarial and deep learning for model training. In both cases, the sample data needs to be annotated. Still, the number of sample data needed to train the models appears to be comparably low. 

On the one hand, the approach based on WGAN, Pix2Pix and U-Net combines three different deep learning architectures and only requires parameter tuning to work for the given dataset. This workflow already offers a high level of automation since it reduces the effort for designing a defect detection system to providing a small sample set of annotated data. Of course, only a well-selected and sufficiently annotated test set allows for serious model validation and testing. 
% On the other hand: human-readible math function
On the other hand, the trigonometric function is transparent, human-readable and stable. Its output can be visualized and tested against tolerance criteria. It also allows to be tuned by setting limiting parameters, such as window size, orientation or line thickness. Also, auditing and requirements testing can be easily performed. However, the mathematical function lacks of flexiblity in terms of domain transfer. In order to design a mathematical model specific to the problem, domain knowledge needs to be collected and translated into abstract dependencies. Thus, it qualifies for applications with a high need for transparency and stability, e. g. security in critical environments.

However the authors stress that they cannot evaluate the whole extent of GAN based DA, but encourage fellow reseachers to explore the application of this approach to other fields and use cases. We expect benefits especially in the field of deep learning, industrial monitoring and neuroevolution. 

\section{Conclusion and Outlook}\label{sec:conclusion}

In this article, we presented image-to-image translation as a means for data augmentation in the context of defect detection on textiles and carbon fiber in particular. Therefore, we discussed related GAN approaches and designed two variations of a novel concept for generating synthetic defects based on sparse labeled data using a \textit{pix2pix} model. 

Within our experiments on six different datasets we showed that a \textit{pix2pix} based approach could substantially improve the pixel-based classification quality of U-Net models when using a problem-specific label generator. In general, the synthetic defects helped to augment the dataset so that segmentation quality improves on sparse data. However, the approach did not clearly outperform regular DA techniques. Still, domain-specific modelling of defect images allows to achieve similiar quality scores. Although WGAN proved inferior to competing techniques, it still helped to support semantic segmentation to a certain degree.

Furthermore, the approach can be used to train neural networks for semantic segmentation on comparably sparse data since GANs manage to generate realistic, yet artifical labels from few samples. Fuzzballs and misaligned fibers serve as a model for industrial camera based surface monitoring in the manufacturing process. The suggested approach has been tested for the given setting but is not limited to textiles. The example was selected in order to demonstrate how the approach behaves for complex detection tasks. The authors deem further experimentation necessary to evaluate the whole potential, i. e. explore and demonstrate the proposed workflow in other domains and for defects that are maybe even less complex. It should be mentioned, that by using the label generator with mathematical models, training data can always be created for a specific use case as for carbon fiber images. Conventional DA only applies very general image transformations without any reference to specific requirements in the application scenario. For that reason, we also suggested WGAN for a higher level of automation. In conclusion, we could reveal great potential for problem-tailored function models but could not verify the same for the WGAN based apporach. The authors, however, suggest to conduct hyperparameter tuning on the GAN models for improvements in future work. The study suggests that semi-supervised training exhibits high relevance for defect detection in industrial applications. Furthermore, it could also be demonstrated that traditional DA did not substantially improve the pixel-based classification. Under the given circumstances the assumption can be made that GAN based augmentation already provides a well-balanced and diverse dataset so that conventional image transformation methods do not add any additional value.

% =======
% Outlook
% =======
Since the current research activities still focus on the adaptation of network models for semantic segmentation and object detection in different areas of industrial image data, we are going to explore the potential for a wide application of GAN training with very little supervision. For this purpose, future research will mainly address the design of networks which allow to design models that are trained with few or no data and still manage to effectively apply industrial monitoring solutions. This will result in problem-specific algorithm development in order to decrease efforts spent on defect simulation and shift it to domain independent methods. We deem the field of neuroevolution in combination with transfer and few-shot learning as promising for industrial applications. Future work will strive for a closer look to hyperparameter optimization in the context of deep learning. Therefore, we plan to extend our concepts with methods from the fields of \textit{Evolutionary} and \textit{Organic Computing} to equip our approach with self-configuring and self-learning properties. The application of evolutionary computation, i.e. genetic algorithms and co-evolution, constitutes another topic of our research agenda. 

% \vfill
\section*{\uppercase{Acknowledgements}}
\noindent The authors would like to thank the Administration of Swabia and the Bavarian Ministry of Economic Affairs and Media, Energy and Technology for funding and support to conduct this research as part of the program \textit{Competence Expansion of Fraunhofer IGCV formerly Fraunhofer Project Group for "Functional Lightweight Design" FIL of ICT}.

\bibliographystyle{apalike}
{\small
\bibliography{main}}

\end{document}